\documentclass[10pt,twocolumn,letterpaper]{article}

\usepackage{iccv}
\usepackage{times}
\usepackage{epsfig}
\usepackage{graphicx}
\usepackage{amsmath}
\usepackage{amssymb}

\usepackage{graphicx}
\usepackage{amsmath}
\usepackage{amssymb}
\usepackage{booktabs}
\usepackage{multirow}
\usepackage{colortbl}
\usepackage{color,xcolor}
\definecolor{mygray}{gray}{.9}

\usepackage{marvosym}  
\usepackage{caption}  
\usepackage{bm} 

\newcommand{\CUT}[1]{}

\usepackage[breaklinks=true,bookmarks=false]{hyperref}

\iccvfinalcopy 

\def\iccvPaperID{****} 

\ificcvfinal\fi

\begin{document}

\title{ReliableSwap: Boosting General Face Swapping Via Reliable Supervision}

\author{Ge Yuan\textsuperscript{1$\dag$}
\qquad
Maomao Li\textsuperscript{2$\dag$}
\qquad
Yong Zhang\textsuperscript{2*}
\qquad
Huicheng Zheng\textsuperscript{1*}\\
\textsuperscript{1}Sun Yat-sen University
\qquad
\textsuperscript{2}Tencent AI Lab\\
}
\maketitle

\let\thefootnote\relax\footnotetext{Work done when Ge Yuan was an intern at Tencent AI Lab.}
\let\thefootnote\relax\footnotetext{$\dag$ Equal contribution.}
\let\thefootnote\relax\footnotetext{$*$ Corresponding authors.}


\begin{abstract}
Almost all advanced face swapping approaches use reconstruction as the proxy task, i.e., supervision only exists when the target and source belong to the same person. Otherwise, lacking pixel-level supervision, these methods struggle for source identity preservation. This paper proposes to construct reliable supervision, dubbed \textbf{cycle triplets}, which serves as the image-level guidance when the source identity differs from the target one during training. Specifically, we use face reenactment and blending techniques to synthesize the swapped face from real images in advance, where the synthetic face preserves source identity and target attributes. However, there may be some artifacts in such a synthetic face. To avoid the potential artifacts and drive the distribution of the network output close to the natural one, we reversely take synthetic images as input while the real face as reliable supervision during the training stage of face swapping. Besides, we empirically find that the existing methods tend to lose lower-face details like face shape and mouth from the source. This paper additionally designs a \textbf{FixerNet}, providing discriminative embeddings of lower faces as an enhancement. Our face swapping framework, named \textbf{ReliableSwap}, can boost the performance of any existing face swapping network with negligible overhead. Extensive experiments demonstrate the efficacy of our ReliableSwap, especially in identity preservation. The project page is \url{https://reliable-swap.github.io/}.
\end{abstract}

\section{Introduction}
Face swapping aims to transfer the identity of a source face into a target one, while maintaining the rest of attributes, \textit{e.g.,} background, light, head pose, and expression.
It has a wide application in the privacy protection~\cite{blanz2004cgf, photorealistic2014accv}, film industry~\cite{comb2020naruniec}, and face forgery detection~\cite{deeepfakedetection2022tpami, twobranch2020eccv}.

    \begin{figure}[t]
      \centering
    \includegraphics[width=1.0\linewidth]{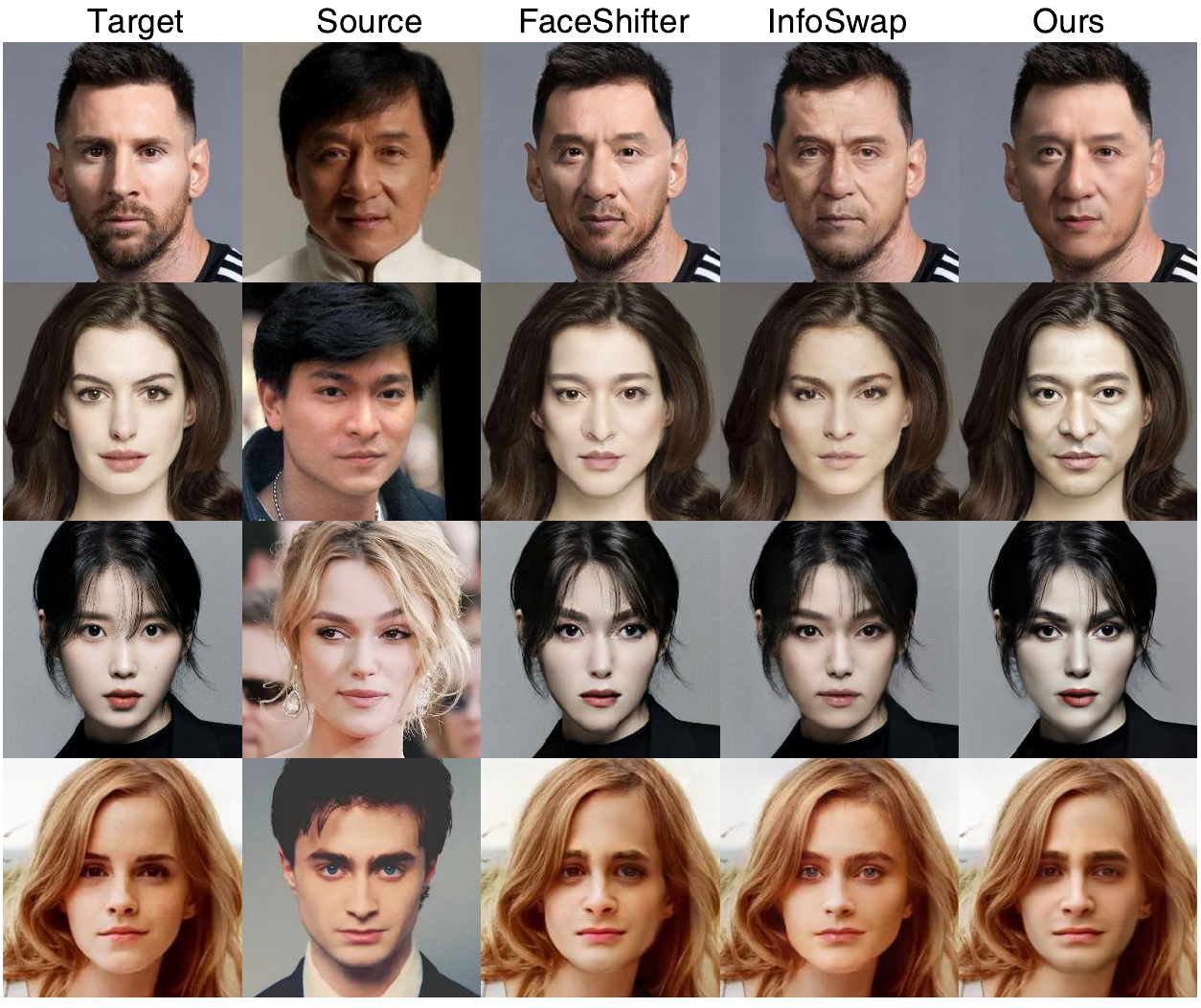}
       \vspace{-0.4cm}
       \caption{Compared with the current state-of-the-art face swapping approaches InfoSwap~\cite{infoswap2021cvpr} and Faceshifter~\cite{faceshifter2020cvpr}, our proposed ReliableSwap achieves better identity preservation from sources. Besides, we discover that the local details of the lower face, such as face shape and mouth, largely affect visual similarity but tend to be neglected by the previous quantitative metrics.} 
       \label{fig:fr_weakness}
       \vspace{-0.2cm}
    \end{figure}

Although fruitful endeavors have been pursued~\cite{fsgan2019iccv, simswap2020mm, faceshifter2020cvpr, infoswap2021cvpr, megafs2021cvpr, hires2022cvpr} on face swapping, existing methods suffer from a common issue: the interpolation identity of the source and target faces. 
That is, rather than keeping the source identity to the maximum, the swapped result resembles neither source nor target, but seems like an interpolated identity between them.
As seen in Fig.~\ref{fig:fr_weakness}, taking two state-of-the-art methods InfoSwap~\cite{infoswap2021cvpr} and Faceshifter~\cite{faceshifter2020cvpr} for illustration,
in terms of overall visual similarity, the swapped and source faces fail to fall into the same identity especially when we additionally make the target face as a reference.
Worse, as shown in the 2nd and 4th rows, the swapped results present inconsistent gender compared with the sources.
Besides, these methods are inclined to lose the local details like mouth and lower face shape even if they achieve high scores on quantitative identity metrics.



A design flaw is responsible for the aforementioned interpolation identity issue. During training, given the target and source of different identities, there is no pixel-wise supervision to guide synthesis in the previous methods~\cite{faceshifter2020cvpr, simswap2020mm, smoothswap2022cvpr}. 
To deal with this, they pick 20\%$\sim$50\% training input pairs and set the source and target to be the same person in each pair.
For these pairs, face swapping can leverage re-construction as the proxy task, and make the target face as the pixel-wise supervision.
Nonetheless, the remaining pairs \CUT{that are with different identities} still lack pixel-level supervision. To handle this, previous methods make efforts to devise sophisticated network architectures~\cite{infoswap2021cvpr, megafs2021cvpr, styleswap2022eccv} or introduce cumbersome priors~\cite{fsgan2019iccv, hififace2021ijcai, rafswap2022cvpr}, but achieving little improvements.

   \begin{table}[t]\small
    \vspace{-0cm}
    \centering
    \begin{tabular}{c|ccccc} 
    \hline
     Changing part       & -  & eyes & nose & mouth & jaw   \\ 
     \hline
    \textit{ID Sim.}$\uparrow$ & 1.00 & 0.76 & 0.90 & 0.91  & 0.95  \\
    \hline
    \end{tabular}
        \caption{FR networks are more sensitive to upper face modification. We change one facial part in turn and keep the others unchanged. Then, we calculate identity similarity (\textit{ID Sim.}) between the corresponding changed faces with the ground truth ones via FR embeddings. Here, we omit the results of identity retrieval (\textit{ID Ret.}), since there is little difference among them.}
        \label{taba:fixernet}
        \vspace{-3mm}
    \end{table}

%

Furthermore, although previous face swapping methods tend to lose details in lower faces, such as the mouth and lower face shape, the widely used identity metrics evaluated through deep face recognition (FR) networks~\cite{cosface2018cvpr, arcface2019cvpr} cannot fully measure such a lost. That is, even though the swapped results appear obviously inconsistent lower face details with the sources, the existing approaches can achieve a high identity retrieval score (\textit{ID Ret.}) and identity similarity (\textit{ID Sim.}) between the source face and the swapped one. Here, we argue that the reason is that the common-used FR networks~\cite{cosface2018cvpr, arcface2019cvpr} are less sensitive to lower face modifications than upper ones~\cite{interpretface2019iccv, from2021tpami}. To validate this, we conduct a pilot experiment. Concretely,  we modify one facial part at a time while remaining the others unchanged and then 
compute the identity similarity \textit{ID Sim.} between the changed faces and the ground-truth ones via FR embedding. 
The results in Tab.~\ref{taba:fixernet} demonstrate that compared with changing the upper face (eyes), changing lower face parts (nose, mouth, jaw) has a much smaller impact on the \textit{ID Sim.}. 
For brevity, we leave the detailed modification process in the Supplementary Material.



\begin{figure}[t]
      \centering
   \includegraphics[width=0.8\linewidth]{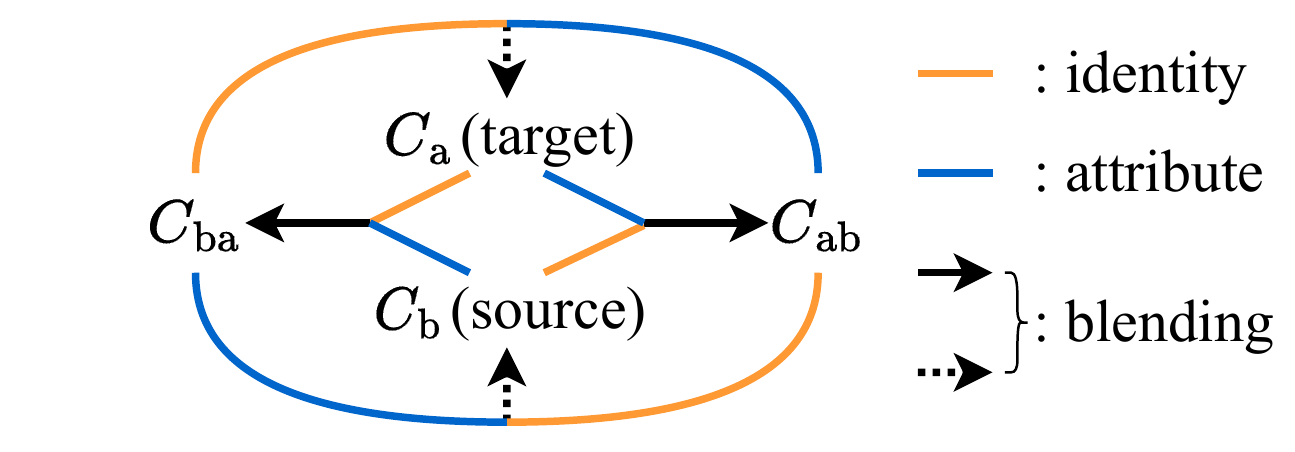}
   \vspace{-1mm}
       \caption{The cycle relationship among the items of cycle triplets.}
       \label{fig:cycle}
       \vspace{-2mm}
    \end{figure}

To deal with the first design flaw, in this work, we construct reliable supervision, named \textbf{cycle triplets}, serving as the image-level guidance for the unsupervised face swapping task.
Specifically, as seen in Fig.~\ref{fig:cycle}, given two real images (the target $C_{\rm{a}}$ and the source $C_{\rm{b}}$), we blend the face of $C_{\rm{b}}$ into $C_{\rm{a}}$ through 
face reenactment~\cite{lia} and multi-band blending~\cite{multiband1983tog}), obtaining the synthesized swapped face $C_{\rm{ab}}$.
These techniques ensure the high-level semantics (identity) are unchanged when pasting a blob of connected pixels (facial regions) from the source $C_{\rm{b}}$ to the target $C_{\rm{a}}$.
Thus, $C_{\rm{ab}}$ inherits identity from the source $C_{\rm{b}}$ and other identity-irrelevant attributes from the target $C_{\rm{a}}$.
Similarly, blending the face of $C_{\rm{a}}$ into $C_{\rm{b}}$ produces another synthesized swapped face $C_{\rm{ba}}$.
As a result, $C_{\rm{ab}}$ preserves the identity from $C_{\rm{b}}$, and $C_{\rm{ba}}$ maintains the attributes from $C_{\rm{b}}$. 
Then, when using the synthesized results $C_{\rm{ba}}$ as the target input and $C_{\rm{ab}}$ as the source one, an ideal face swapping model would output $C_{\rm{b}}$ as the result, which forms cycle relationship.
In this paper, we name the image triplet $\{C_{\rm{ba}}, C_{\rm{ab}}, C_{\rm{b}}\}$ as a cycle triplet, where another cycle triplet $\{C_{\rm{ab}}, C_{\rm{ba}}, C_{\rm{a}}\}$ can also be constructed in the same fashion. 
Given that both of $C_{\rm{ab}}$ and $C_{\rm{ba}}$ are with some artifacts inevitably, we use synthetic faces $C_{\rm{ba}}$ and $C_{\rm{ab}}$ as input, while a real image $C_{\rm{b}}$ as the reliable supervision.
A similar situation can also be generalized when we can take the target $C_{\rm{a}}$ as supervision.
In this way, the proposed cycle triplets would encourage the distribution of network output close to natural images and avoid potential artifacts.


Second, to enhance the lower face details, we extra propose a FixerNet, which can be easily inserted into the existing face swapping methods with little overhead. 
Specifically, our FixerNet embeds the discriminative features of the lower face as a supplement to the identity embedding of the whole face. 
Feeding such discriminative embedding additionally to the existing face swapping networks can guide these models to generate faces with more consistent lower face patterns and fix those potentially lost details, which motivates its name FixerNet.
Besides, to quantitatively demonstrate the effectiveness of our FixerNet, we propose two new metrics: lower-face identity retrieval (\textit{L Ret.}) and lower-face identity similarity (\textit{L Sim.}) to evaluate the performance of face swapping methods on lower-face details.
In a nutshell, the contributions of this work can be summarized as:
\vspace{-0.2cm}
\begin{itemize}
    \item {We propose to construct cycle triplets as reliable supervisions to boost general face swapping methods, where we take the synthetic images as input while the real images as guidance.}
    \vspace{-0.2cm}
    \item{We present a FixerNet to remedy lower face details, where we additionally propose two new metrics to evaluate the performance on lower face identity.}
    \item 
    \vspace{-0.2cm}
    {Our face swapping framework, dubbed ReliableSwap,
    can be incorporated with any existing face swapping methods flexibly. Based on the Faceshifter~\cite{faceshifter2020cvpr}, our ReliableSwap achieves new state-of-the-art face-swapping performance.}
\end{itemize}


\section{Related Work}
\vspace{-0.1cm}
\subsection{Approaches Based on Image-Level Blending}
Early face swapping methods~\cite{blanz2004cgf, bitouk2008TOG} use traditional computer graphic (CG) approaches~\cite{multiband1983tog, poisson2003siggraph} to blend two faces at image level.
Recently, FastSwap~\cite{korshunova2017ICCV} leverages a multi-scale convolutional neural network for image-to-image translation.
Improved on~\cite{nirkin2018fg}, FSGAN~\cite{fsgan2019iccv} reenacts source faces by a GAN~\cite{gan2014nips}, freeing the requirements of sophisticated 3D priors.
Naruniec et al.~\cite{comb2020naruniec} propose a high-resolution encoder-decoder network, but each target demands a tailored decoder.
Famous open-source algorithms DeepFakes~\cite{deepfakes2021github} and DeepFaceLab~\cite{deepfacelab2020arxiv} provide full pipelines for face swapping.
These methods follow the same idea, i.e., blending the faces with similar pose and expression by traditional CG methods.
However, they suffer from the unnatural swapped result and obvious artifacts occurring on the blending boundaries.

\CUT{AOT[] formulates appearance transfer as an optimal transport[] problem and relight the swapped face based on 3D mesh and normals.
HifiFace~\cite{hififace2021ijcai} combines 3DMM vectors and face recognition embeddings as the source identities.}

\vspace{-0.1cm}
\subsection{Feature-Based Methods}
\noindent{\textbf{Extracting or Disentangling Features.}}
With 3DMM~\cite{deep3d2019cvprw}, some face swapping methods~\cite{nirkin2018fg, drl2019cvpr, aot2020nips, hififace2021ijcai} disentangle shape and texture features from the source for subsequent latent blending.
Following GANs~\cite{gan2014nips}, a group of methods~\cite{rsgan2018arXiv, fsnet2018ACCV, ipgan2018cvpr} disentangle identity features through adversarial learning. 
Besides, inspired by mutual information, Gao et al.~\cite{infoswap2021cvpr} present information bottlenecks for compact features.
SmoothSwap~\cite{smoothswap2022cvpr} trains an identity embedder via contrastive learning~\cite{contrastive2020icml} for a smoother feature space.

Recently, for more disentangled features, various approaches~\cite{megafs2021cvpr,rafswap2022cvpr,hires2022cvpr}
assume the input distribution of a pre-trained StyleGAN generator~\cite{stylegan2019cvpr, stylegan22020cvpr} as their prior distributions.
Specifically, MegaFS~\cite{megafs2021cvpr} swaps the multi-level features of the source and target in the W++ space~\cite{stylegan22020cvpr}.
FaceInpainter~\cite{faceinpainter2021cvpr} adapts identity swapping to various domains.
Based on Transformer~\cite{transformer2017nips}, RAFSwap~\cite{rafswap2022cvpr} projects face parsing information into identity features.
HiRes~\cite{hires2022cvpr} modulates the pose and expression of the source according to facial landmarks.
Although StyleGANs can disentangle features of different semantics, how to control features to fit the desired visual patterns remains unsolved, limiting its applications on face swapping.

\noindent{\textbf{Fusing Features of Identity and Attributes.}}
Another line of work studies to fuse features better.
\cite{localin2020accv} proposes a feature blending scheme for synthetic faces.
\cite{simswap2020mm, faceshifter2020cvpr, megafs2021cvpr, styleswap2022eccv} blend the features using the predicted latent masks. 
SimSwap~\cite{simswap2020mm} injects source identity features into the reconstruction of the target face.
FaceShifter~\cite{faceshifter2020cvpr} fuses identity and multi-level attribute features in a decoder.
To compress the model, MobileFS~\cite{mobeilefaceswap2022aaai} uses depthwise separable convolution~\cite{mobilenetv22018CVPR} and dynamic neural network~\cite{dynamic2016NIPS} to adjust the student's weights according to the teacher.
Based on a StyleGAN2~\cite{stylegan22020cvpr} trained from scratch, StyleSwap~\cite{styleswap2022eccv} concatenates attribute features to StyleGAN2 layers; however, it requires fine-tuning during testing.

Different from the previous methods, we propose to boost general face swapping by constructing cycle triplets as reliable supervision, confronting the unsupervised challenges. Furthermore, for more comprehensive facial patterns, we design a FixerNet to compensate for the lost details like lower face shape and mouth.




  \begin{figure}[t]
       \centering
       \includegraphics[width=1.0\linewidth]{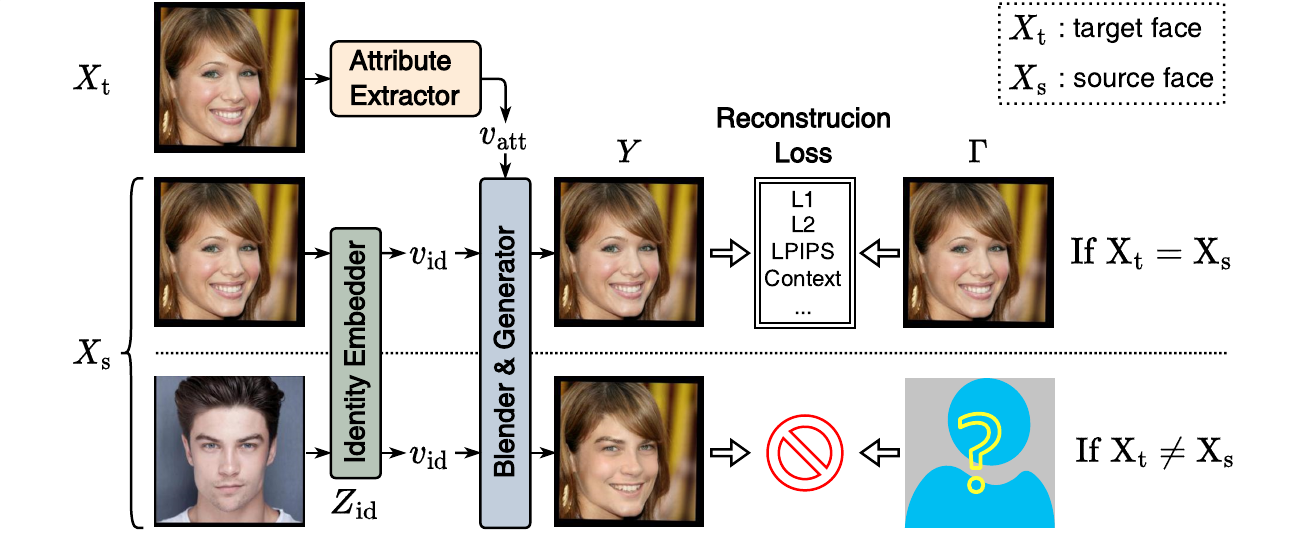}
       \caption{The typical face swapping training process.}
       \label{fig:typical_pipeline}
       \vspace{-3mm}
    \end{figure}

   \begin{figure*}[ht]
       \centering
       \includegraphics[width=0.96\linewidth]{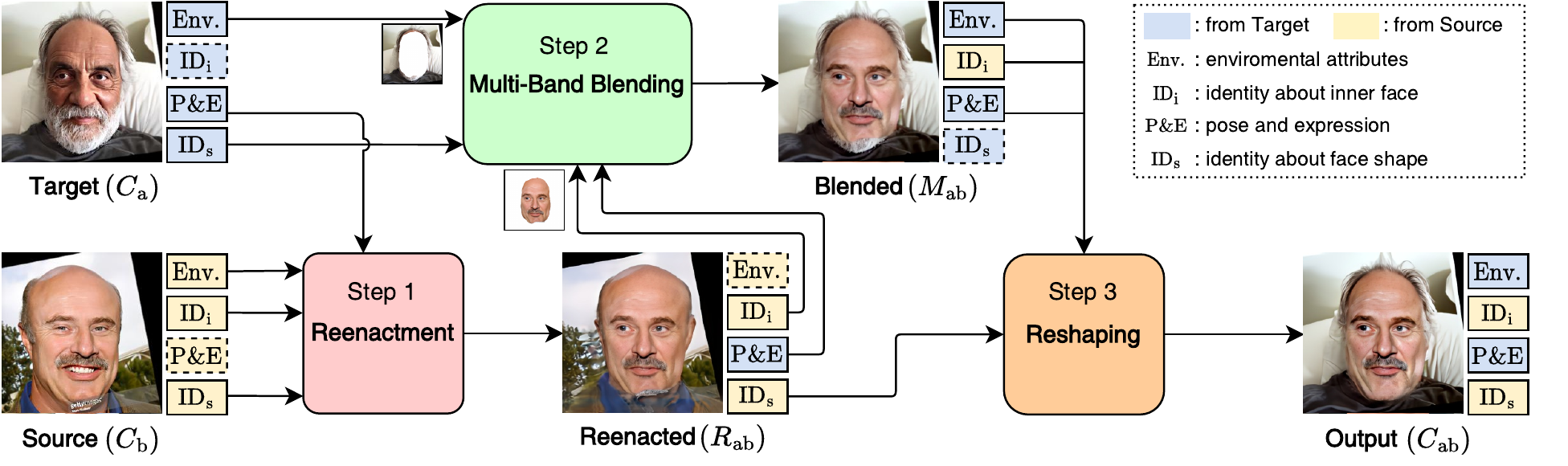}
       \caption{The pipeline of synthesizing fake images and obtaining naive triplets, which consists of three steps. The Reenactment step first transfers pose and expression from the target $C_{\rm{a}}$, leading to the reenacted face $R_{\rm{ab}}$. Then, the Multi-Band Blending step blends inner faces from the reenacted source $R_{\rm{ab}}$ to the target $C_{\rm{a}}$, bringing a coarse swapped face image $M_{\rm{ab}}$. Last, the Reshaping step remedies potential inconsistency of face shape and outputs the synthetic swapped result $C_{\rm{ab}}$.}
       \label{fig:triplet_overall}
    \end{figure*}



\vspace{-0.1cm}
\section{Method}
\subsection{Preliminaries}
As illustrated in Fig.~\ref{fig:typical_pipeline}, we first review the typical training framework of previous face swapping methods.
The identity features $v_{\rm{id}}$ of the source $X_{\rm{s}}$ are extracted by a pre-trained FR network $Z_{\rm{id}}$, and the other identity-irrelevant attributes $v_{\rm{att}}$ of the target $X_{\rm{t}}$ are obtained by an attribute extractor.
Then, a feature blender merges $v_{\rm{id}}$ and $v_{\rm{att}}$, followed by a generator predicting the swapped face $Y$. During training, the reconstruction loss is used to penalize the similarity between $Y$ and an image reference $\Gamma$ ($=X_{\rm{t}}$) if and only if $X_{\rm{t}}=X_{\rm{s}}$.
This case takes up 20\%$\sim$50\% of the training samples.
However, when $X_{\rm{t}} \neq X_{\rm{s}}$, there is no image-level reference to guide the generation of $Y$, where re-construction cannot be used as the proxy anymore.
In this way, the lack of pixel-wise supervision increases the uncertainty of synthesized results, potentially weakening the preservation of source identity.
To deal with this, we propose to construct reliable training supervision in advance, which encourages the swapped result consistent with the source identity to the maximum, yielding a high-fidelity face swapping.

\subsection{Synthesizing to Obtain Naive Triplets}

    \begin{figure}[ht]
       \centering
       \includegraphics[width=1.0\linewidth]{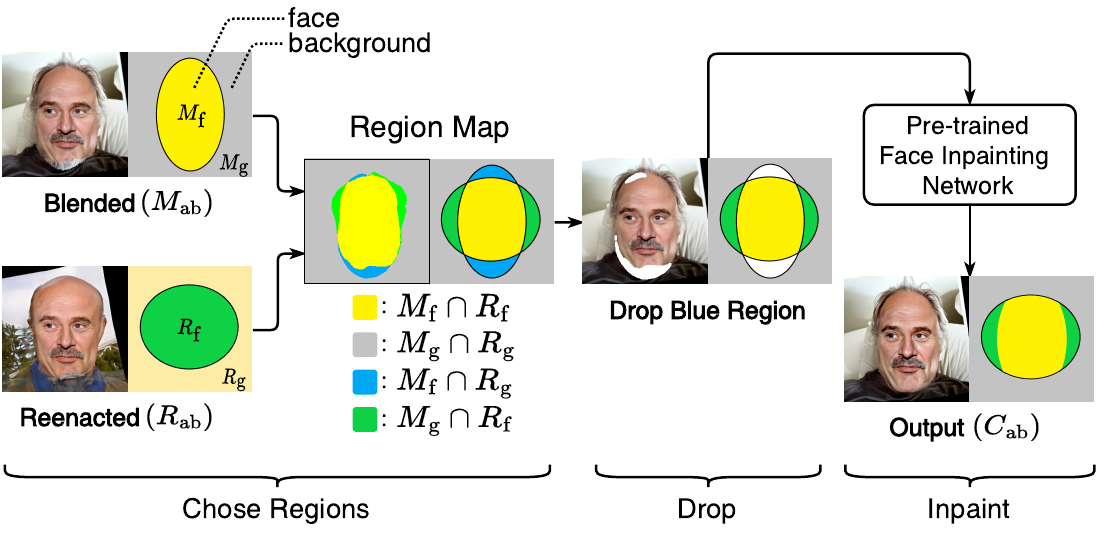}
       \caption{Segmentation-based dropping and inpainting in the Reshaping step.}
       \label{fig:triplet_shape}
       \vspace{-3mm}
    \end{figure}


Formally, as illustrated in Fig.~\ref{fig:triplet_overall},
we define that a face image consists of four components: environment ($\rm{Env.}$) including foreground, background, and light; pose and expression ($\rm{P\&E}$); inner face ($\rm{ID_i}$), like eyes, nose, and mouth; face shape ($\rm{ID_s}$), respectively. 
Given $C_{\rm{a}}$ as the target and $C_{\rm{b}}$ as the source, we first synthesize the swapped faces $C_{\rm{ab}}$, which maintains the inner face $\rm{ID_i}$ and face shape $\rm{ID_s}$ with the source $C_{\rm{b}}$, and keep the environment $\rm{Env.}$ and the pose and expression $\rm{P\&E}$ from the target $C_{\rm{a}}$.

Then, the pipeline of synthesizing the naive triplet $\{C_{\rm{a}},C_{\rm{b}},C_{\rm{ab}}\}$  can be formulated as three steps: Reenactment, Multi-Band Blending, and Reshaping.
Note that the separated description of $\rm{ID_i}$ and $\rm{ID_s}$ is to demonstrate that our synthesized swapped face $C_{\rm{ab}}$ would give extra consideration to maintaining the face shape from the source $C_{\rm{b}}$ with the Reshaping step.


First, reenactment aims to transfer the pose and expression of a driving image to a source image while keeping the source identity unchanged. In this paper, to encourage the synthesized swapped face $C_{\rm{ab}}$ to be with the same pose and expression as the target face $C_{\rm{a}}$, the proposed Reenactment step modulates $\rm{P\&E}$ of $C_{\rm{b}}$ towards $C_{\rm{a}}$ with the reenactment model LIA~\cite{lia}, obtaining the reenacted face $R_{\rm{ab}}$.

Then, we use a face parsing~\cite{dmlcsr2022cvpr} model to estimate facial segmentation masks for the reenacted face $R_{\rm{ab}}$ and the target $C_{\rm{a}}$, with which
we blend the $R_{\rm{ab}}$ into $C_{\rm{a}}$ via Multi-Band Blending~\cite{multiband1983tog}, leading to a coarse swapped face image $M_{\rm{ab}}$. 
Here, the environment attribute $\rm{Env.}$ and the pose and expression $\rm{P\&E}$ in $M_{\rm{ab}}$ are well preserved from the target $C_{\rm{a}}$, while the inner face $\rm{ID_i}$ are consistent with $R_{\rm{ab}}$ as well as the source $C_{\rm{b}}$.


Next, although the coarse swapped face image $M_{\rm{ab}}$ have the same inner face $\rm{ID_i}$ with the reenacted face $R_{\rm{ab}}$, there is no clear constrain for the face shape $\rm{ID_s}$ in $M_{\rm{ab}}$. Sometimes, when the target face $C_{\rm{a}}$ is fatter than the reenacted one $R_{\rm{ab}}$, the face shape of $M_{\rm{ab}}$ would be consistent with that of the target, which deviates from our goal of maintaining both inner face and face shape with the reenacted face $R_{\rm{ab}}$ as well as the source $C_{\rm{b}}$.
To deal with such face-shape inconsistency, as detailed in Fig.~\ref{fig:triplet_shape}, we propose to refine the coarse swapped image $M_{\rm{ab}}$ with the Reshaping Step, which is based on the facial segmentation maps.

Formally, let $M_{\rm{f}}$ denotes the foreground facial region of the coarse swapped image $M_{\rm{ab}}$, and $M_{\rm{g}}$ denotes the background. 
Similarly, we denote $R_{\rm{f}}$ as the facial region of the reenacted image $R_{\rm{ab}}$ and $R_{\rm{g}}$ as the background.
Then, we mix up the regions of $M_{\rm{ab}}$ and $R_{\rm{ab}}$, obtaining a {Region Map}.
The yellow region $=M_{\rm{f}} \cap R_{\rm{f}}$ is the facial-region intersection of $M_{\rm{ab}}$ and $R_{\rm{ab}}$.
The gray region $=M_{\rm{g}} \cap R_{\rm{g}}$ represents the background intersection of $M_{\rm{ab}}$ and $R_{\rm{ab}}$.
The green region $=M_{\rm{g}} \cap R_{\rm{f}}$ denotes the bulge of the reenacted face $R_{\rm{ab}}$.
All these three regions do not evolve with the Reshaping process and would be kept in the synthesized image $C_{\rm{ab}}$.
The blue region $=M_{\rm{f}} \cap R_{\rm{g}}$ indicates the bulge of the coarse swapped image $M_{\rm{ab}}$. To maintain the face shape of the reenacted face $R_{\rm{ab}}$, we drop the blue region and inpaint it with the background $M_{\rm{g}}$ using a  pre-trained face inpainting network~\cite{deepfill2019iccv}. 


Analogously, we can use $C_{\rm{a}}$ as the source and $C_{\rm{b}}$ as the target to generate another synthetic swapped face $C_{\rm{ba}}$, thus obtaining two naive triplets $\{C_{\rm{a}},C_{\rm{b}},C_{\rm{ab}}\}$ and $\{C_{\rm{b}},C_{\rm{a}},C_{\rm{ba}}\}$.

\subsection{Training with Cycle Triplets}
As seen in Tab.~\ref{tab:input_output}, we list training inputs $X_{\rm{t}}$ and $X_{\rm{s}}$, predictions $Y$, and possible reference image $\Gamma$ during training \CUT{of} a face swapping network.
Since there may be an unnatural appearance in the synthesized swapped faces $C_{\rm{ab}}$ and $C_{\rm{ba}}$, directly using naive triplets as $\{ X_{\rm{t}},X_{\rm{s}},\Gamma \}$ can be suboptimal, which would make the distribution of the network output $Y$ far from a natural one.

In this paper, we inversely take the synthesized swapped faces $C_{\rm{ba}}$ and $C_{\rm{ab}}$ as input, while real $C_{\rm{b}}$ as the reliable supervision. As illustrated in Fig.~\ref{fig:cycle}, taking as input the attribute features of $C_{\rm{ba}}$ and the identity ones of $C_{\rm{ab}}$, an ideal face swapping network should predict a swapped result identical to the source $C_{\rm{b}}$. That is, we can construct a \textbf{cycle triplet} 
$\{ C_{\rm{ba}}, C_{\rm{ab}}, C_{\rm{b}}\}$ by painlessly rotating the element order in naive triplets. The another cycle triplet $\{C_{\rm{ab}}, C_{\rm{ba}}, C_{\rm{a}}\}$ can be generated with a similar fashion.


    \begin{table}[t]
    \centering
    \begin{tabular}{l|ll|l|l}
    \toprule[1pt]
                      & $X_{\rm{t}}$ & $X_{\rm{s}}$ & $Y$ & $\Gamma$  \\ 
    \midrule
    \multirow{2}{*}{vanilla training samples} &  $I_{\rm{t}}$  & $I_{\rm{s}}$  & $Y_{\rm{t,s}}$  & \textcolor{red}{None}  \\
                      &  $I_{\rm{t}}$  &  $I_{\rm{t}}$  & $Y_{\rm{t,t}}$  & $I_{\rm{t}}$    \\ 
    \midrule
    \multirow{2}{*}{naive triplets (ours)} 
        &  $C_{\rm{a}}$  & $C_{\rm{b}}$    & $Y_{\rm{a,b}}$  &  $C_{\rm{ab}}$  \\
        &  $C_{\rm{b}}$  &  $C_{\rm{a}}$  & $Y_{\rm{b,a}}$  & $C_{\rm{ba}}$    \\
    \midrule
        \rowcolor{mygray}
        &   $C_{\rm{ab}}$  & $C_{\rm{ba}}$  &  $Y_{\rm{ab,ba}}$ &  $C_{\rm{a}}$ \\
        \rowcolor{mygray}
        \multirow{-2}{*}{ cycle triplets (ours)} &  $C_{\rm{ba}}$  &  $C_{\rm{ab}}$  &  $Y_{\rm{ba,ab}}$ & $C_{\rm{b}}$ \\
    \bottomrule[1pt]                  
    \end{tabular}
    \caption{Training inputs and outputs of face swapping. The rows in \colorbox{mygray}{gray} indicate our cycle triplets joining the training.}
        \label{tab:input_output}
    \end{table}

The proposed cycle triplets can remedy the absence of reference (row 1 in Tab.~\ref{tab:input_output}) when $X_{\rm{t}}$ and $X_{\rm{s}}$ belong to different identities for existing face swapping approaches.
Following the commonly-used reconstruction loss~\cite{ipgan2018cvpr}, we calculate a cycle-triplet loss $\mathcal{L}_{\rm{ct}}$ between $Y$ and $\Gamma$ when $X_{\rm{t}}=C_{\rm{ba}}$ (or $C_{\rm{ab}}$) and $X_{\rm{s}}=C_{\rm{ab}}$ (or $C_{\rm{ba}}$).
The cycle-triplet loss $\mathcal{L}_{\rm{ct}}$ contains a pixel-wise consistent loss $\mathcal{L}^{\rm{ct}}_{\rm{pixel}}$, a Learned Perceptual Image Path Similarity (LPIPS) loss $\mathcal{L}^{\rm{ct}}_{\rm{LPIPS}}$~\cite{lpips2018cvpr}, and an identity loss $\mathcal{L}^{\rm{ct}}_{\rm{id}}$, which can be expressed as:
\begin{gather}
\label{eq:loss_tri}
    \mathcal{L}^{\rm{ct}}_{\rm{pixel}} = \Vert Y - \Gamma \Vert_1, \text{ if } \Gamma \in \{C_{\rm{a}},C_{\rm{b}}\},  \\
    \mathcal{L}^{\rm{ct}}_{\rm{LPIPS}} = \Vert \rm{VGG}(Y) - \rm{VGG}(\Gamma) \Vert_1, \text{ if } \Gamma \in \{C_{\rm{a}},C_{\rm{b}}\}, \\
    \mathcal{L}^{\rm{ct}}_{\rm{id}} = 1 - \cos(Z_{\rm{id}}(Y), Z_{\rm{id}}(\Gamma)), \text{ if } \Gamma \in \{C_{\rm{a}},C_{\rm{b}}\},
\end{gather}
where $\Vert \cdot \Vert_1$ denotes $L_1$ loss, $\rm{VGG}(\cdot)$ represents a VGGNet~\cite{vggnet2014arxiv} extracting perceptual features, $Z_{\rm{id}}$ denotes the identity extractor which is usually a pre-trained FR network, and $\cos(\cdot, \cdot)$ indicates the cosine similarity between two embeddings
obtained from face recognition networks.
Thus, the total cycle triplet loss is calculated as:
\begin{gather}
\label{eq:loss_tri}
    \mathcal{L}_{\rm{ct}} = \lambda^{\rm{ct}}_1 \mathcal{L}^{\rm{ct}}_{\rm{pixel}} + \lambda^{\rm{ct}}_2 \mathcal{L}^{\rm{ct}}_{\rm{LPIPS}} + \lambda^{\rm{ct}}_3 \mathcal{L}^{\rm{ct}}_{\rm{id}},
\end{gather}
where $\lambda^{\rm{ct}}_1$, $\lambda^{\rm{ct}}_2$, and $\lambda^{\rm{ct}}_3$ are the hyper-parameters that control the trade-off between these three terms.

To constrain the domain of training inputs close to the natural distribution, we mix up the cycle triplets with vanilla training samples. In essence,
the synthetic images ($C_{\rm{ab}}$ and $C_{\rm{ba}}$) in cycle triplets can be treated as data augmentation, potentially improving the robustness of the model.

\subsection{FixerNet}

\begin{figure}[t]
       \centering
       \includegraphics[width=0.9\linewidth]{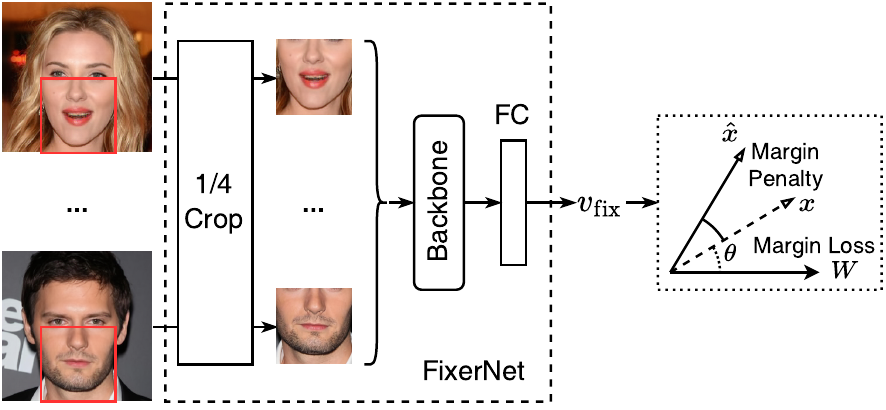}
       \caption{Training stage of the proposed FixerNet.}
       \label{fig:fixernet}
    \end{figure}

\noindent{\textbf{The Details of the FixerNet.}}
Recall that previous methods tend to lose lower face details (e.g., lower face shape and mouth), to remedy this, we further present a FixerNet as an additional identity extractor.
For discriminative lower-face embeddings, we train the FixerNet on a large face dataset MS1M~\cite{ms1m2026eccv} with identity annotations.
As shown in Fig.~\ref{fig:fixernet}, we use the detected and aligned faces as the training samples.
Then we crop the middle parts of a lower half face, where the cropped size ($56\times 56$) is a quarter of the holistic aligned image ($112\times 112$). A deep network backbone like ResNet~\cite{deep2016cvpr} takes as input these cropped samples. Consequently, the fully-connected (FC) layer embeds a latent feature $v_{\rm{fix}}$ under the supervision of a margined softmax loss~\cite{arcface2019cvpr}.
The embedded $v_{\rm{fix}}$ represents the identity-discriminative features of the lower face.


FixerNet can be painlessly plugged into existing face swapping networks.
During training, we use the pre-trained FixerNet $Z_{\rm{fix}}$ to extract $v_{\rm{fix}}$ from $X_{\rm{s}}$ and concatenate it with the vanilla identity embedding $v_{\rm{id}}$ by $v_{\rm{full}} = [v_{\rm{id}};v_{\rm{fix}}]$.
Here, $[\cdot;\cdot]$ indicates the concatenation of two tensors at the last dimension.
Our $v_{\rm{full}}$ substitutes the vanilla $v_{\rm{id}}$ as the source identity feature input of the Blender and Generator in Fig.~\ref{fig:typical_pipeline}.
Besides, we present a Fixer loss to penalize the lower face similarity between $X_{\rm{s}}$ and $Y$:
\begin{equation}
    \mathcal{L}^{\rm{fix}}_{\rm{s}} = 1 - \cos(Z_{\rm{fix}}(X_{\rm{s}}), Z_{\rm{fix}}(Y)).
\end{equation}

Furthermore, Fixer loss can cooperate with cycle triplets, where the lower face information provided by supervision $\rm{\Gamma}$ in cycle triplet can be used to guide the generation of $Y$:
\begin{equation}
    \mathcal{L}^{\rm{fix}}_{\rm{\Gamma}} = 1 - \cos(Z_{\rm{fix}}(\Gamma), Z_{\rm{fix}}(Y)), \text{ if } \Gamma \in \{C_{\rm{a}}, C_{\rm{b}}\}.
\end{equation}

The total Fixer loss function is formulated as:
\begin{equation}
    \mathcal{L}_{\rm{fix}} = \lambda^{\rm{fix}}_1 \mathcal{L}^{\rm{fix}}_{\rm{s}} + \lambda^{\rm{fix}}_2 \mathcal{L}^{\rm{fix}}_{\rm{\Gamma}},
\end{equation}
where $\lambda^{\rm{fix}}_1$ and $\lambda^{\rm{fix}}_2$are loss weights.

\noindent{\textbf{New Metrics For Lower-face Performance.}}
The current face swapping community~\cite{faceshifter2020cvpr, infoswap2021cvpr, hires2022cvpr} leverages the retrieval score (\textit{ID Ret.}) and cosine similarity (\textit{ID Sim.}) of embeddings extracted by a FR network to measure source identity preservation of the swapped results.
Such a identity extractor is instantiated as CosFace~\cite{cosface2018cvpr} during the face swapping training procedure ($Z_{\rm{id}}$ in Fig.~\ref{fig:typical_pipeline}), while as ArcFace~\cite{arcface2019cvpr} during evaluation. However, as shown in the pilot experiment in the Introduction section,
these two metrics cannot fully evaluate the lower-face details of face swapping results since FR embeddings are more sensitive to upper face modification. 

To deal with this, we follow the above design principle and propose two corresponding new metrics: lower-face identity retrieval \textit{L Ret.} and lower-face identity similarity \textit{L Sim.}.
Specifically, we use the different dataset, backbone, and loss function with those of training FixerNet to obtain a new pre-trained network denoted as $L_{\rm{net}}$.
Then, we can calculate \textit{L Ret.} and \textit{L Sim.} by extracting discriminate embeddings of lower faces by the obtained $L_{\rm{net}}$. Here, our \textit{L Ret.} and \textit{L Sim.} can be regarded as a complement for the existing \textit{ID Ret.} and \textit{ID Sim.}.



\section{Experiments}

\subsection{Experimental Setup}
\noindent{\textbf{Face Swapping Datasets.}}
We use VGGFace2~\cite{cao2018vggface2} as the training dataset, which contains 3.3M face images. We crop and align these images following FFHQ~\cite{stylegan2019cvpr}.
After calculating the IQA scores~\cite{serfiq2020cvpr}, we filter the top 1.5M images and resize them to 256$\times$256. 
FaceForensics++~\cite{faceforensics2019iccv} and CelebA-HQ~\cite{karras2017progressive} datasets are used to evaluate the methods.

\noindent{\textbf{The Settings of Cycle Triplets and FixerNet.}}
Before training face swapping networks, we construct 600k cycle triplets offline, whose number is 40\% of vanilla training samples.
Then we use an IQA filter~\cite{serfiq2020cvpr} to drop the images with low quality.
The backbone, dataset, and identification loss of FixerNet are ResNet-50~\cite{deep2016cvpr}, MS1M~\cite{ms1m2026eccv},  and ArcFace Loss~\cite{arcface2019cvpr}, while those of $L_{\rm{net}}$ are IResNet-50~\cite{arcface2019cvpr}, CASIA-WebFace~\cite{casia2014arxiv}, and CosFace Loss~\cite{cosface2018cvpr}.

    \begin{figure}[t]
      \centering
       \includegraphics[width=1.0\linewidth]{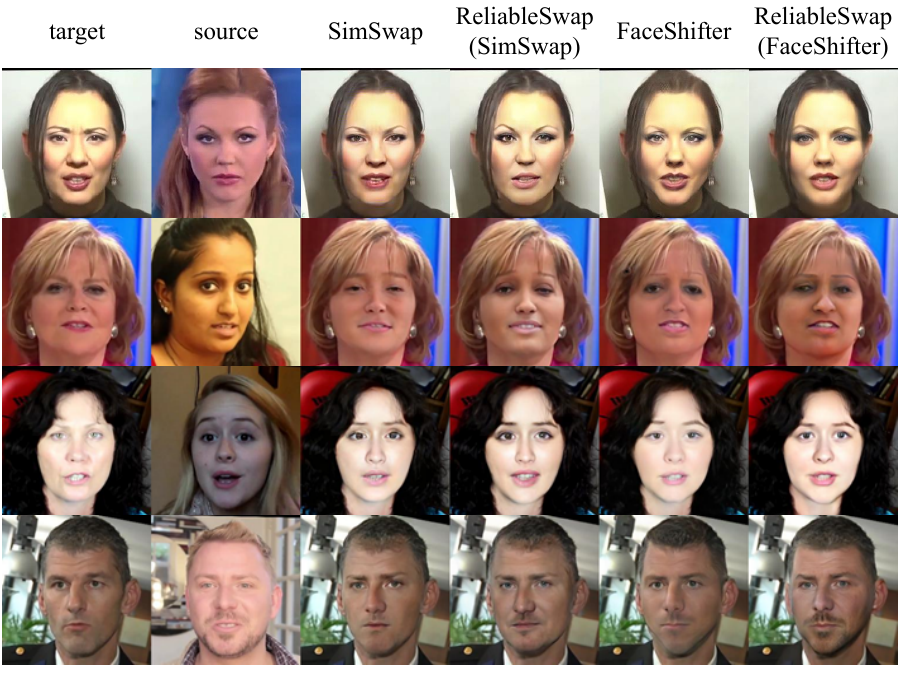}
       \vspace{-5mm}
       \caption{Qualitative comparison between two baseline methods SimSwap, FaceShifter and our ReliableSwap (\emph{w/} SimSwap) as well as ReliableSwap (\emph{w/} FaceShifter).}
       \label{fig:ffplus}
       \vspace{-3mm}
    \end{figure}

    \begin{figure}[ht]
      \centering
       \includegraphics[width=0.75\linewidth]{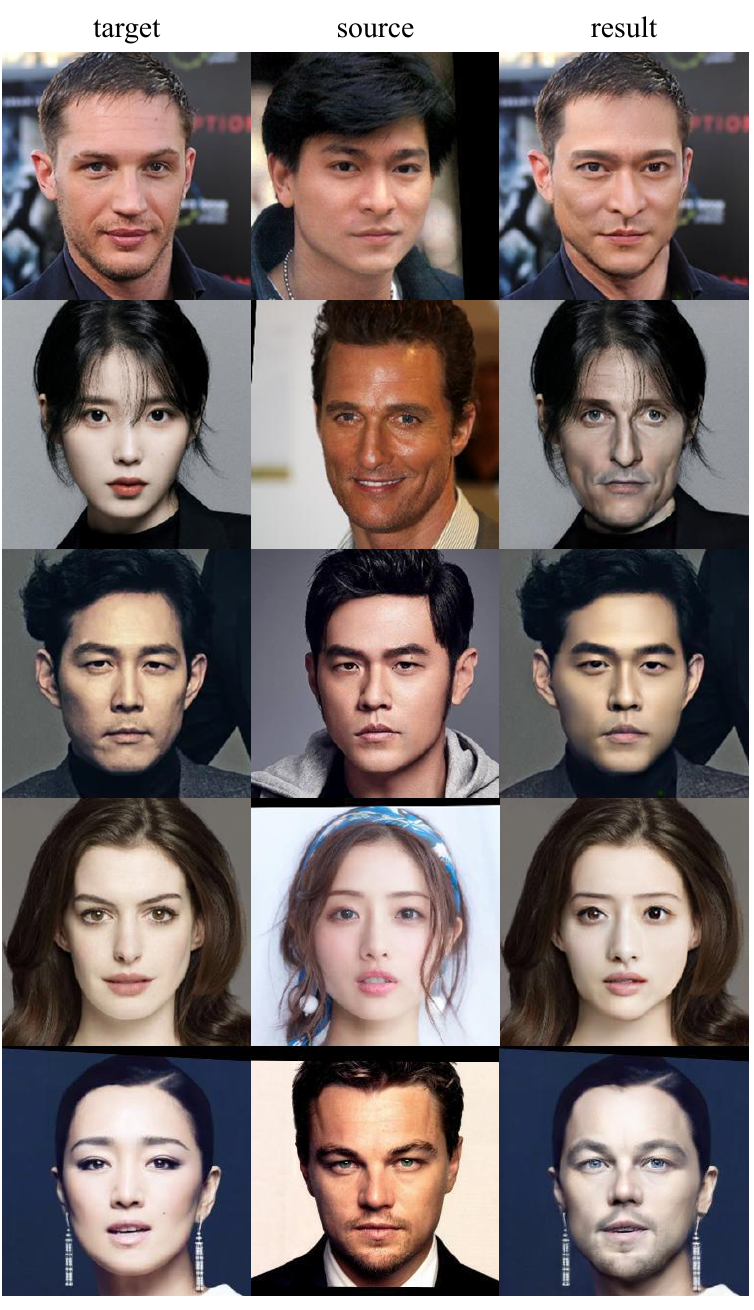}
       \vspace{-1mm}
       \caption{Face swapping results on images collected from web.}
       \label{fig:web}
    \end{figure}

    \begin{table}
    \centering
    \footnotesize
    \begin{tabular}{lllll} 
    \toprule[1pt]
    Method             &\textit{ID Ret.}$\uparrow$ &\textit{L Ret.}$\uparrow$& \textit{Pose}$\downarrow$  & \textit{Exp.}$\downarrow$  \\
    \midrule 
    DeepFakes~\cite{deepfakes2021github}          &   88.39   &  10.43  &   4.46   &   3.33    \\
    MegaFS~\cite{megafs2021cvpr}             &   90.83  &  33.89  &   2.64   &   \textbf{2.73}    \\
    InfoSwap~\cite{infoswap2021cvpr}           &   90.09   &  47.96 &   2.21   &   3.12    \\
    HiRes ~\cite{hires2022cvpr}             &   90.05  &  19.63  &   2.58   &   3.16    \\ 
    \midrule
    SimSwap~\cite{simswap2020mm}*           &   88.34   &  41.37 &   1.69   &   \textbf{2.86}    \\
    Ours (\emph{w/} \cite{simswap2020mm})     &  91.91    &  78.62 &  \textbf{1.62}    &   2.87    \\
    FaceShifter~\cite{faceshifter2020cvpr}*       &   90.02   &  51.23  &   2.33   &   3.09    \\
    \rowcolor{mygray} Ours (\emph{w/} \cite{faceshifter2020cvpr}) &   \textbf{93.44}  &  \textbf{82.99}  &  2.25    &   3.09    \\
    \bottomrule[1pt]
    \end{tabular}
    \caption{Quantitative evaluation results on the FaceForensics++ dataset on the \textit{ID Ret.}, \textit{L Ret.}, \textit{Pose}, and \textit{Exp.}, where ``*'' denotes we reproduce the results.}
    \label{tab:quanti_F++}
    \vspace{-1mm}
    \end{table}

\noindent{\textbf{Training Details.}}
We choose two SOTA open-source face swapping algorithms SimSwap~\cite{simswap2020mm} and FaceShifter~\cite{faceshifter2020cvpr} as the baselines of our ReliableSwap. 
For fair comparisons, we apply the same training recipes including batch size, training steps, and learning rate of the Adam optimizer~\cite{adam}.
Our cycle triplet loss and Fixer loss are added to the original baseline losses, whose increase $\sim4\%$ training time.
Please refer to the Supplementary Materials for more detailed experimental settings and model complexity comparisons.


    \begin{figure*}[ht]
      \centering
       \includegraphics[width=0.95\linewidth]{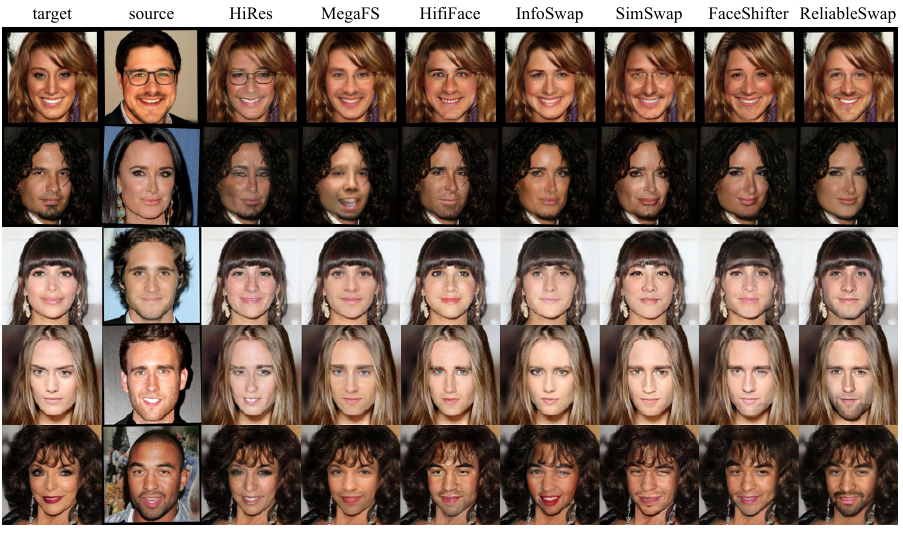}
       \vspace{-3mm}
       \caption{Qualitative face swapping results on the CelebA-HQ dataset.}
       \label{fig:celeb}
       \vspace{-3mm}
    \end{figure*}


\subsection{Comparison with SOTA Methods}
\noindent{\textbf{Qualitative Comparison.}}
Following the evaluation protocol of~\cite{faceshifter2020cvpr}, we compare our ReliableSwap with two different baselines in Fig.~\ref{fig:ffplus}. 
We show various scenarios, where the input target and source images have a large gap in face shape, mouth, expression, pose, and light condition.
The results demonstrate that our ReliableSwap preserves more identity details.
Furthermore, we evaluate ReliableSwap on wild celebrity faces collected from movies and Internet in Fig.~\ref{fig:web}.
Benefiting from the reliable supervision provided by cycle triplets and lower facial details kept through FixerNet, our results preserve high-fidelity source identity, including nose, mouth, and face shape.

Then, in Fig.~\ref{fig:celeb}, we compare several competitive methods HiRes~\cite{hires2022cvpr}, MegaFS~\cite{megafs2021cvpr},
Hififace~\cite{hififace2021ijcai}, InfoSwap~\cite{infoswap2021cvpr}, SimSwap~\cite{simswap2020mm}, and FaceShifter~\cite{faceshifter2020cvpr} with our ReliableSwap (\emph{w/} FaceShifter) on the CelebA-HQ dataset.
Specifically, we sample five pairs with obvious variants in gender, skin color, pose, and expression.
Our ReliableSwap outperforms others on source identity preservation, as well as global similarity and local details.

    \begin{table}
    \centering
    \footnotesize
    \begin{tabular}{llllll} 
    \toprule[1pt]
    Method             &\textit{ID Sim.}$\uparrow$ & \textit{L Sim.}$\uparrow$ & \textit{Pose}$\downarrow$ & \textit{Exp.}$\downarrow$ & \textit{FID}$\downarrow$  \\ 
    \midrule
    MegaFS~\cite{megafs2021cvpr}             &   0.3173   &  0.3740 &   4.20   & \textbf{2.65}    &   10.35   \\
    InfoSwap~\cite{infoswap2021cvpr}           &   0.3843  &  0.4046  &   \textbf{2.40}   &  3.00    &  \textbf{6.45}   \\
    HiRes~\cite{hires2022cvpr}              &   0.2922   & 0.2993  &   3.12   &  3.15    &   7.46   \\ 
    \midrule
    FaceShifter~\cite{faceshifter2020cvpr}    &   0.4335   & 0.4152  &   2.78   & 3.12    &   9.00   \\
    \rowcolor{mygray} Ours (\emph{w/} \cite{faceshifter2020cvpr}) &\textbf{0.4731}  &  \textbf{0.5227}  &   2.64   &   3.12   &   6.90   \\
    \bottomrule[1pt]
    \end{tabular}
    \caption{Quantitative evaluation results on the CelebA-HQ dataset in terms of \textit{ID Sim.}, \textit{L Sim.}, \textit{Pose}, \textit{Exp.}, and \textit{FID}.}
    \label{fig:cele_hq}
    \vspace{-3mm}
    \end{table}

    \begin{table}[t]
    \small
    \centering
    \begin{tabular}{lll} 
    \toprule[1pt]
    Method  &Identity $\uparrow$   &Attributes $\uparrow$ \\
    \midrule
    MegaFs~\cite{megafs2021cvpr}  &  1.81  &    5.07    \\
    InfoSwap~\cite{infoswap2021cvpr}   & 15.42  &    19.31    \\
    HiRes~\cite{hires2022cvpr}  & 6.24  &    11.46      \\
 Faceshifter~\cite{faceshifter2020cvpr} &  15.76 &  \textbf{32.88} \\
    \rowcolor{mygray} ReliableSwap (\emph{w/} \cite{faceshifter2020cvpr})  &  \textbf{60.77} &   31.28   \\
    \bottomrule[1pt] 
    \end{tabular}
    \caption{Human study results (\%), where we show the averaged selection percentages of each method.}
    \label{userstudy}
    \vspace{-3mm}
    \end{table}

\noindent{\textbf{Quantitative Comparison.}}    
In Tab.~\ref{tab:quanti_F++}, we follow the FaceForensics++ evaluation protocol~\cite{faceshifter2020cvpr} to display the quantitative performances on identity retrieval (\textit{ID Ret.} and \textit{L Ret.}), head pose errors (\textit{Pose}), and expression errors (\textit{Exp.}).
Specifically, we first sample 10 frames from each video and process them by MTCNN~\cite{mtcnn}, obtaining 10K aligned faces.
Then we take these 10K faces as target inputs, whereas the corresponding source inputs are the same as those in the FaceShifter.

As for \textit{ID Ret.}, we use CosFace~\cite{cosface2018cvpr} to extract identity embedding with dimension 512 and retrieve the closest face by cosine similarity.
To evaluate pose and expression,  we use HopeNet~\cite{pose} as the pose estimator and Deep3D~\cite{chaudhuri2019joint} as the expression feature extractor. 
Then we measure the $L_2$ distances between these features extracted from the swapped result and the corresponding target inputs.
The results in Tab.~\ref{tab:quanti_F++} show that our ReliableSwap improves the identity consistency on SimSwap and FaceShifter.
Besides, ours based on FaceShifter achieves the highest \textit{ID Ret.} and  \textit{L Ret.} and ours based on SimSwap are with best \textit{Pose} and and comparable \textit{Exp.}, which demonstrates the efficacy of the proposed method.

Following RAFSwap~\cite{rafswap2022cvpr}, we randomly sample 100K image pairs from CelebA-HQ as the evaluation benchmark.
We report identity similarity (\textit{ID Sim.}) and (\textit{L Sim.}), pose errors (\textit{Pose}), expression errors (\textit{Exp.}), and FID~\cite{fid2017nips} (\textit{FID}) in Tab.~\ref{fig:cele_hq}.
Our ReliableSwap achieves the best identity preservation, and comparable \textit{Pose}, \textit{Exp.}, and \textit{FID}.  

\noindent{\textbf{Human Study.}}
We conduct a human study to compare our ReliableSwap with the SOTA methods.
Corresponding to two key objectives of face swapping, we ask users to choose:
\textit{a) the one most resembling the source face, 
b) the one keeping the most identity-irrelevant attributes with the target face.}
For each user, we randomly sample 30 pairs from the images used in the above qualitative comparison.
We report the selected ratios based on the answers of 100 users in Tab.~\ref{userstudy}, where the results demonstrate our method surpasses all other methods on identity similarity and achieves competitive attribute preservation.

\subsection{Analysis of ReliableSwap}

    \begin{figure}[t]
       \centering
       \includegraphics[width=0.9\linewidth]{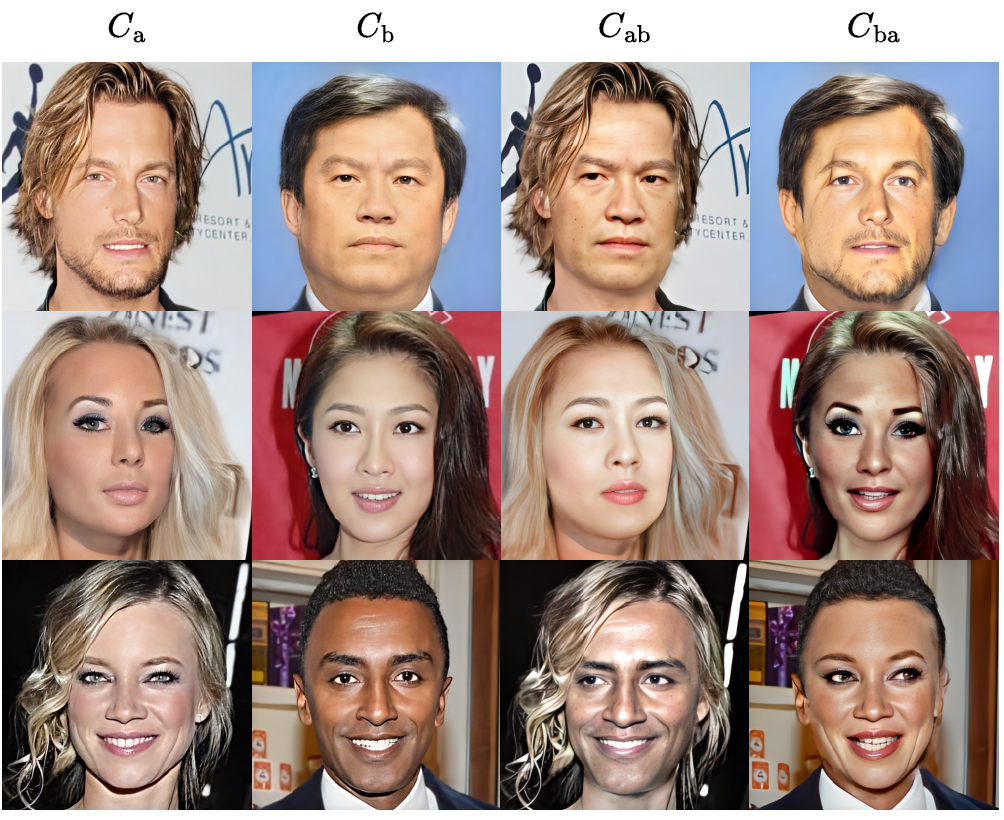}
       \caption{Examples of cycle triplets.}
       \label{fig:lia_good}
    \end{figure}

    \begin{table}[t]
    \centering
    \small
    \begin{tabular}{lllll} 
    \toprule[1pt]
    Method              & \textit{ID Ret.}$\uparrow$ & \textit{L Ret.}$\uparrow$ & \textit{Pose}$\downarrow$ & \textit{Exp.} $\downarrow$ \\ 
    \midrule
    FaceShifter ~\cite{faceshifter2020cvpr}        &  90.02  &  51.23   & 2.33   &  \textbf{3.09}     \\
    FixerNet            &  90.11   &  77.44  &  2.31    &  3.10     \\
    200k cycle triplets &  92.22   &  58.21  &  2.30    &  3.10     \\
    600k cycle triplets &  93.08   &  63.21  &  2.29    &  3.09     \\
    ReliableSwap           &\textbf{93.44}   &  \textbf{82.99}  &  \textbf{2.25}    &  \textbf{3.09}     \\
    \bottomrule[1pt] 
    \end{tabular}
    \caption{Quantitative ablation study on FaceForensics++ using\textit{ID Ret.}, \textit{L Ret.}, \textit{Pose}, \textit{Exp.}.}
    \label{tab:ablation}
    \vspace{-0.15cm}
    \end{table}

\noindent{\textbf{The Examples of Cycle Triplets.}}
As seen in Fig.~\ref{fig:lia_good}, we provide some examples of cycle triplets, where $C_{\rm{ab}}$ (or $C_{\rm{ba}}$) keeps the true identity of $C_{\rm{b}}$ (or $C_{\rm{a}}$) but with an unnatural appearance.
That is why we use cycle triplets instead of navie ones during training face swapping networks.

\begin{figure}[t]
      \centering
       \includegraphics[width=0.95\linewidth]{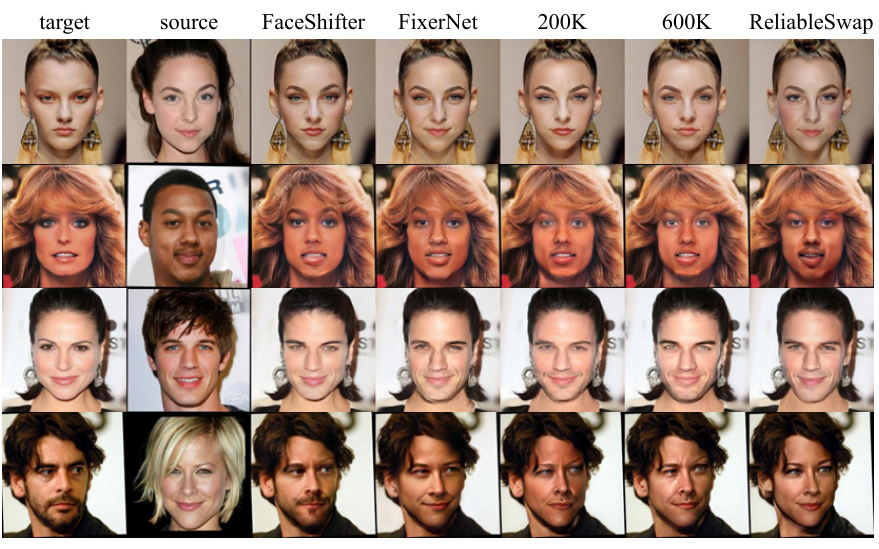}
       \caption{Qualitative comparison of different ablation variants.}
       \label{fig:ablation}
       \vspace{-3mm}
    \end{figure}

\noindent{\textbf{The Number of Cycle Triplets.}}
To verify the efficacy of the proposed cycle triplets, we train three models with different numbers of cycle triplets (0, 200K, and 600K) and compare their qualitative and quantitative evaluation results.
Here, we use the vanilla FaceShifter as the baseline, where the proposed FixerNet is disabled.
As shown in Fig.~\ref{fig:ablation}, the results of column 3 are with a clear
interpolation identity issue.
In contrast, our two methods with 200K (column 5) and 600K (column 6) cycle triplets generate more identity-consistent faces, whereas the one with more cycle triplets preserves more source identity information.
Furthermore, we provide a quantitative comparison in Tab.~\ref{tab:ablation}. It can be seen that the model with 200K cycle triplets improves \textit{ID Ret.} and \textit{L Ret.} by 2.20 and 6.98 over the FaceShifter baseline, respectively.
Increasing the number of cycle triplets to 600K can further improve \textit{ID Ret.} and \textit{L Ret.}.

\noindent{\textbf{FixerNet.}}
To validate that the proposed FixerNet can maintain local facial details of the lower face, we insert it into the vanilla Faceshifter baseline and Faceshifter trained with 600K cycle triplets, respectively. As shown in Fig.~\ref{fig:ablation}, 
no matter in the vanilla Faceshifter (column 3) or the one trained with 600K cycle triplets (column 6), our FixerNet can boost their performance on the local face details (see column 4 and column 7).
The face shape and mouth in those results which are with FixerNet can be transferred better from the source face.
 
Note that since FR embeddings are less insensitive to the changes on the lower face (see the Introduction section), the improvement on \textit{ID Ret.} brought by FixerNet in Tab.~\ref{tab:ablation} seems to be marginal. As contrast, such an improvement on the proposed \textit{ID Ret.} can be 26.21.

\section{Conclusion}

In this paper, we propose a general face swapping framework, named ReliableSwap, which can boost the performance of any existing face swapping network with negligible overhead.
Our ReliableSwap tackles the interpolated identity preservation problem by constructing cycle triplets to provide reliable image-level supervision.
Specifically, we first synthesize naive triplets via traditional computer graphic algorithms, preserving true identity information.
Then based on the cycle relationship among real and synthetic images, we construct cycle triplets using real images as training supervision.
Further, we present a FixerNet to compensate for the loss of lower face details.
Our ReliableSwap achieves state-of-the-art performance on the FaceForensics++ and CelebA-HQ datasets and other wild faces, which demonstrates the superiority of our method.


{\small
\bibliographystyle{ieee_fullname}
\bibliography{egbib}
}

\appendix
\clearpage 
\noindent{\Large\bf Appendix}
\vspace{5pt}

This Supplementary Material includes seven parts, which are: broader impact of our main paper (Section~\ref{broader}), 
face modification process of our pilot experiment in the Introduction (Section~\ref{pilot}), more implementation details of our ReliableSwap with different baselines and model complexity (Section~\ref{imple} and Section~\ref{complexity}), more analysis of naive triplets (Section~\ref{naive}),
more visualization results (Section~\ref{results}), 
additional experiments on SimSwap~\cite{simswap2020mm} baseline (Section~\ref{simswap}),
comparisions with StyleFace~\cite{luo2022styleface} (Section~\ref{styleface}),
and the demo description of video face swapping (Section~\ref{video}).

\section{Broader Impact}
\label{broader}
Face swapping algorithms provide possibilities for immoral behaviours, including identity theft, disinformation attacks, and celebrity pornography. To avoid abuse, it is meaningful to follow the latest face swapping approaches and study more powerful forgery detection methods based on more reliable synthetic swapped samples. Our ReliableSwap shows the state-of-the-art ability to preserve source identity and target attributes, helping people know the threats of face swapping. We will share the results of ReliableSwap to promote the healthy development of the forgery detection community.

\section{Detailed Setups of Pilot Experiment}
\label{pilot}

    \begin{figure}[h]
      \centering
       \includegraphics[width=1.0\linewidth]{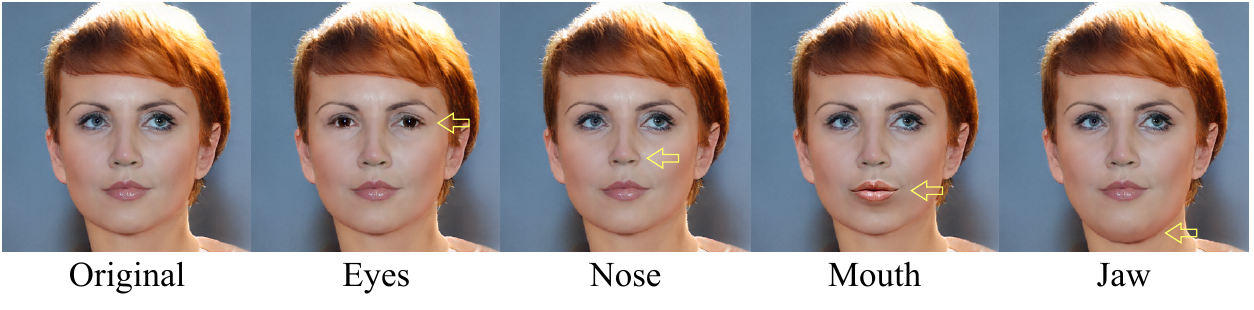}
       \caption{Examples of the corresponding modified faces, where we change one facial part at a time.}
       \label{fig:supp_e4s}
    \end{figure}

Recall that in the Introduction of our main paper, we conduct a pilot experiment to validate that common-used face recognition (FR) networks~\cite{cosface2018cvpr, arcface2019cvpr} are less sensitive to lower face modifications than upper ones~\cite{interpretface2019iccv, from2021tpami}. Specifically, we first use a face editing approach E4S~\cite{e4s2022arxiv} to modify one facial part at a time while remaining the rest parts unchanged. Fig.~\ref{fig:supp_e4s} shows the corresponding examples of modifying eyes, nose, mouth, and jaw for a given face. We randomly choose 1,000 original faces from CelebA-HQ~\cite{karras2017progressive} and obtain \textit{four} kinds of synthetic faces by modifying different parts in turn.
Then we use a widely used FR net ArcFace~\cite{arcface2019cvpr} to extract feature embeddings from 1,000 original and 4$\times$1,000 synthetic faces.
By calculating the average cosine similarity (ID Sim.) between the embeddings of synthetic faces and those of the corresponding original faces, we can compare the FR net's sensitivity to different facial parts.
The experimental results in Tab. 1 demonstrate that the FR net is more sensitive to upper face (eyes) than lower face parts (nose, mouth, jaw).

    \begin{table}[t]
    \footnotesize
    \centering
    \begin{tabular}{l|lll|ll}
    \toprule[1pt]
    Method               & LFW & CFP & AgeDB & Params & FLOPs     \\ 
    \midrule
    ArcFace-100~\cite{arcface2019cvpr}  & 99.77 & 98.27 &  98.28 & 65.16M   & 12.15G    \\
    \rowcolor{mygray} FixerNet         & 99.40 & 95.83 &   95.40   & 27.70M     & 1.31G  \\
    \midrule
    CosFace-50~\cite{cosface2018cvpr} & 99.11 & 94.38 & 91.70 & 49.73M & 6.90G \\ 
    \rowcolor{mygray} $L_{\rm{net}}$    & 98.47  & 92.00 &  89.40  & 31.79M     & 2.06G  \\ 
    \bottomrule[1pt]
    \end{tabular}
    \caption{Comparison of face recognition accuracy and model complexity between widely used pre-trained models in face swapping and our proposed networks.}
    \label{tab:fixernet_flops}
    \end{table}

    \begin{figure*}[h]
      \centering
       \includegraphics[width=1.0\linewidth]{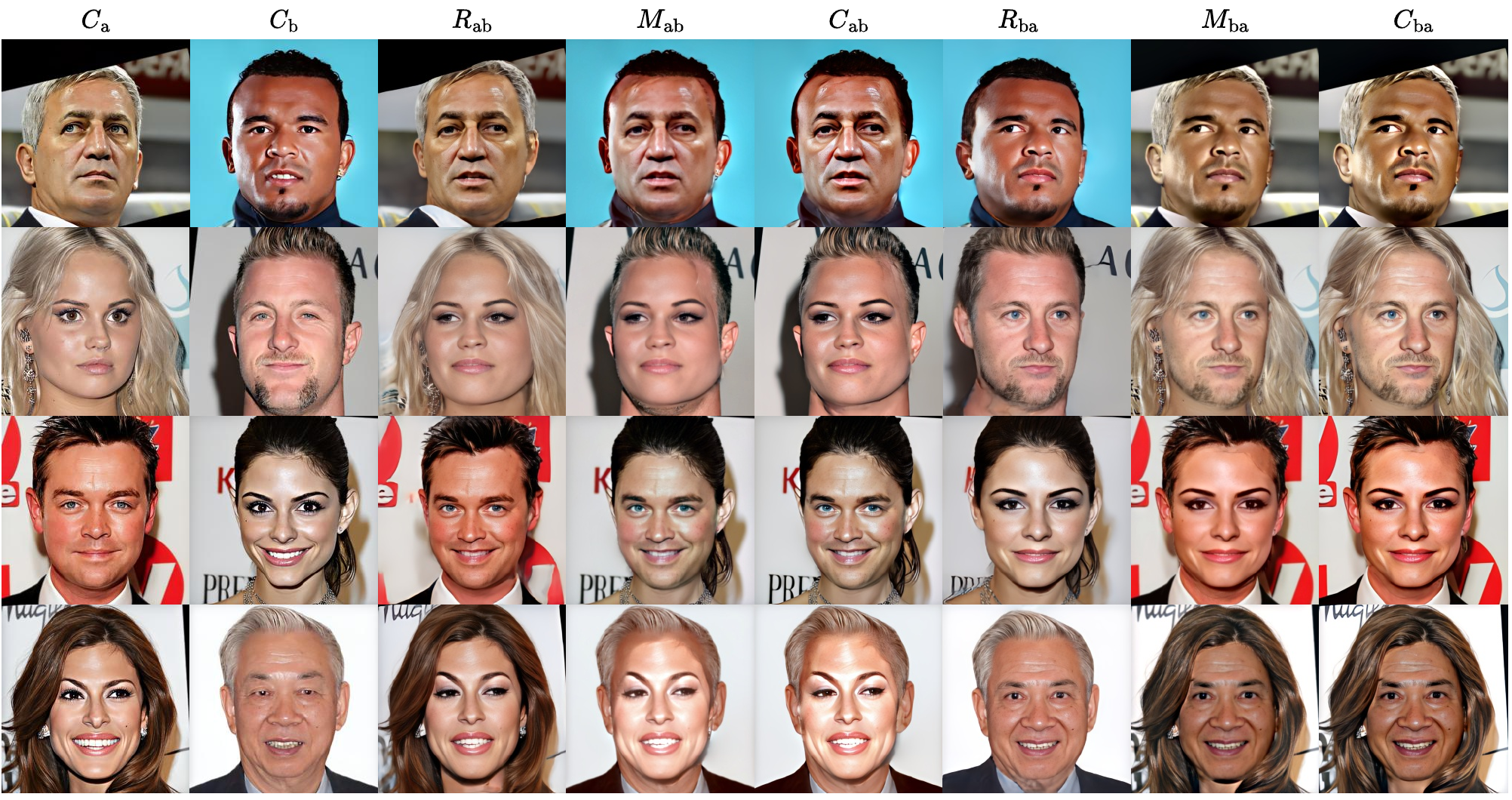}
       \caption{Qualitative results of synthesizing naive triplets.}
       \label{fig:supp_naive}
    \end{figure*}

\section{Additional Implementation Details}
\label{imple}

\noindent{\textbf{Details of Training with Cycle Triplets.}}
For 600k cycle triplets, we use an IQA filter~\cite{serfiq2020cvpr} to drop about 450k low-quality triplets where the images' IQA scores decrease over 0.4 after the synthesizing process.
The fake images of the remaining 150k cycle triplets are mixed with 1,500k vanilla training faces from VGGFace2~\cite{cao2018vggface2} for training face swapping models. 

\noindent{\textbf{Training Details for ReliableSwap (\emph{w/} FaceShifter).}}
For ReliableSwap using FaceShifter~\cite{faceshifter2020cvpr} as the baseline, we set $\lambda^{\rm{ct}}_1$, $\lambda^{\rm{ct}}_2$, $\lambda^{\rm{ct}}_3$, $\lambda^{\rm{fix}}_1$, and $\lambda^{\rm{fix}}_2$ as 1, 5, 10, 1, and 2, separately.
The learning rate of Adam optimizer~\cite{adam} is set to $0.0001$, with hyper-parameters $\beta_1=0$ and $\beta_2=0.999$.

\noindent{\textbf{Training Details for ReliableSwap (\emph{w/} SimSwap).}} 
When using SimSwap~\cite{simswap2020mm} as the baseline for ReliableSwap, we set $\lambda^{\rm{ct}}_1$, $\lambda^{\rm{ct}}_2$, $\lambda^{\rm{ct}}_3$, $\lambda^{\rm{fix}}_1$, and $\lambda^{\rm{fix}}_2$ as 0.5, 5, 10, 0.5, and 0.5, respectively. Besides, 
Adam optimizer~\cite{adam} with learning rate $=0.0004$, $\beta_1=0$ and $\beta_2=0.99$ is used for training.

\section{Model Complexity}
\label{complexity}

\noindent{\textbf{FixerNet and $\bm{L_{\rm{net}}}$.}}
To demonstrate that our FixerNet and $L_{\rm{net}}$ are identity-discriminative on lower face, we evaluate their performance on three face benchmarks: LFW~\cite{05_huang2008labeled}, CFP-CP~\cite{sengupta2016frontal}, and AgeDB-30~\cite{moschoglou2017agedb}, as shown in Tab.~\ref{tab:fixernet_flops}.
We also list the results of the identity embedder ArcFace-100~\cite{arcface2019cvpr} and the identity evaluator CosFace-50~\cite{cosface2018cvpr}, both of which are widely used by existing face swapping approaches~\cite{simswap2020mm, faceshifter2020cvpr, infoswap2021cvpr, hires2022cvpr}.
Comparing with these two models, our FixerNet and $L_{\rm{net}}$ achieve comparable accuracy despite they receive only lower face information, validating their discriminative ability.
Because $L_{\rm{net}}$ is trained on a much smaller dataset (with 0.5M images) comparing with FixerNet (with 4.8M images), $L_{\rm{net}}$ shows lower accuracy even if it has larger Params and FLOPs.
Besides, both the parameters and FLOPs of FixerNet are much less than those of ArcFace-100, which means integrating FixerNet into the existing face swapping methods brings little overhead.

\noindent{\textbf{ReliableSwap.}}
We construct cycle triplets offline before the training of face swapping. 
The total training steps of ReliableSwap are consistent with the corresponding baseline.
Therefore, the potential additional training cost brought by using cycle triplets only comes from the extra loss calculation which accounts for a small fraction of the whole forward and backward propagation computing.
That is, integrating cycle triplets into training samples would bring little impacts on the total training time.
In our ReliableSwap, only the FixerNet slightly increases the model complexity and affects the inference speed.
Tab.~\ref{tab:comparison_flops} lists the model complexity of our ReliableSwap and the other state-of-the-art methods.
The results demonstrate that compared with the both baselines, our method increase around $20\%$ parameters and $6\%$ FLOPs.
Tested on NVIDIA A100 GPU, the FPS of our  ReliableSwap slips about only 2$\sim$3.

    \begin{table}[t]
    \centering
    \small
    \begin{tabular}{l|lll} 
    \toprule[1pt]
    Method               & Params & FLOPs & FPS    \\ 
    \midrule
    SimSwap~\cite{simswap2020mm}  & {120.21M}     & 75.22G  & {19.79}  \\
    FaceShifter~\cite{faceshifter2020cvpr}  & 249.50M     & {47.66G}  & 17.35  \\
    MegaFS~\cite{megafs2021cvpr}      & 321.50M     & 49.67G  & 7.69   \\
    InfoSwap~\cite{infoswap2021cvpr}   & 251.06M     & 374.95G & 2.40   \\ 
    \midrule
    \rowcolor{mygray} Ours (\emph{w/} SimSwap)     & 147.91M     & 76.53G  & 16.91  \\
    \rowcolor{mygray} Ours (\emph{w/} FaceShifter) & 277.20M     & 48.97G  & 14.22  \\
    \bottomrule[1pt]
    \end{tabular}
    \caption{The comparison of model complexity. \CUT{FPS is measured on NVIDIA A100.}}
    \label{tab:comparison_flops}
    \end{table}

    \begin{table}[t]
    \small
    \vspace{-0.cm}
    \centering
    \begin{tabular}{lcccc} 
    \toprule[1pt]
    Step    & \textit{ID Sim.}$\uparrow$ & \textit{Pose}$\downarrow$ & \textit{Exp.}$\downarrow$ & \textit{FID}$\downarrow$  \\ 
    \midrule
    start (source)    & 1.0000                     & 7.12                      & 3.92                      & 2.95                      \\
    s$_1$       & 0.6765                     & 4.77                      & 2.52                      & 7.28                      \\
    s$_1$+s$_2$    & 0.5382                     & 4.58                      & 2.74                      & 17.45                     \\
    s$_1$+s$_2$+s$_3$ & 0.5213                     & 4.69                      & 2.76                      & 16.11                     \\
    \midrule
     FaceShifter~\cite{faceshifter2020cvpr}        &   0.4587   &  3.01  &  3.26    &  9.32  \\
    \bottomrule[1pt]
    \end{tabular}
        \caption{Intermediate results during synthesizing triplets on the VGGFace2. To make the changing degree of these metrics easily understood, we provide the results of FaceShifter~\cite{faceshifter2020cvpr} here for reference.}
        \label{taba:naive}
    \end{table}

\section{Analysis of Naive Triplets}
\label{naive}

The pipeline of synthesizing naive triplets consists of three steps: Reenactment ($\rm{s_1}$), Multi-Band Blending ($\rm{s_2}$), and Reshaping ($\rm{s_3}$).
To show the performance of  each step, we quantitatively and qualitatively evaluate these intermediate results during synthesizing triplets in Tab.~\ref{taba:naive} and Fig.~\ref{fig:supp_naive}.


Tab.~\ref{taba:naive} shows the quantitative changes during synthesizing triplets with VGGFace2~\cite{cao2018vggface2}, where the \textit{ID Sim.} is measured between the corresponding result and the source face while the \textit{Pose} and  \textit{Exp.} is calculated between the corresponding result and the target one.
The step $\rm{s}_1$ modulates the pose and expression of the source to approach the target, which allows our synthesis results to maintain the pose and expression of the target. 
In contrast, step $\rm{s}_2$ and $\rm{s}_3$ rarely change pose and expression.
Comparing with the FaceShifter, the synthesized triplets preserve identity well (high \textit{ID Sim.}) but underperform on target attributes consistency (high \textit{Pose}) and natural quality (high \textit{FID}).
The pose error mainly comes from the step $\rm{s_1}$, where the face reenactment model~\cite{lia} cannot precisely transfer the pose.
The step $\rm{s_2}$ increase the \textit{FID} from 7.28 to 17.45 (higher means worse), which corresponds with the fact that Multi-Band Blending can produce blended results with artifacts and unnatural appearance.

The corresponding qualitative comparison among the results after each step
are showed in Fig.~\ref{fig:supp_naive}, where the reenacted faces $R_{\rm{ab}}$ and $R_{\rm{ba}}$ are output by $\rm{s_1}$, the coarsely blended faces $M_{\rm{ab}}$ and $M_{\rm{ba}}$ are output by $\rm{s_2}$, and the reshaped results $C_{\rm{ab}}$ and $C_{\rm{ab}}$ are finally output by $\rm{s_3}$.
The results $M_{\rm{ab}}$ and $M_{\rm{ba}}$ appear much more unnatural compared with $R_{\rm{ab}}$ and $R_{\rm{ba}}$, which are consistent with the largely increased \textit{FID} after the step $\rm{s_2}$ in Tab.~\ref{taba:naive}.
The final synthesized results $C_{\rm{ab}}$ and $C_{\rm{ba}}$ have obvious unnatural appearance but preserve the target attributes and true source identity of inner face and face shape contour.


\section{Additional Qualitative Results}
\label{results}

    \begin{figure*}[t]
      \centering
       \includegraphics[width=1.0\linewidth]{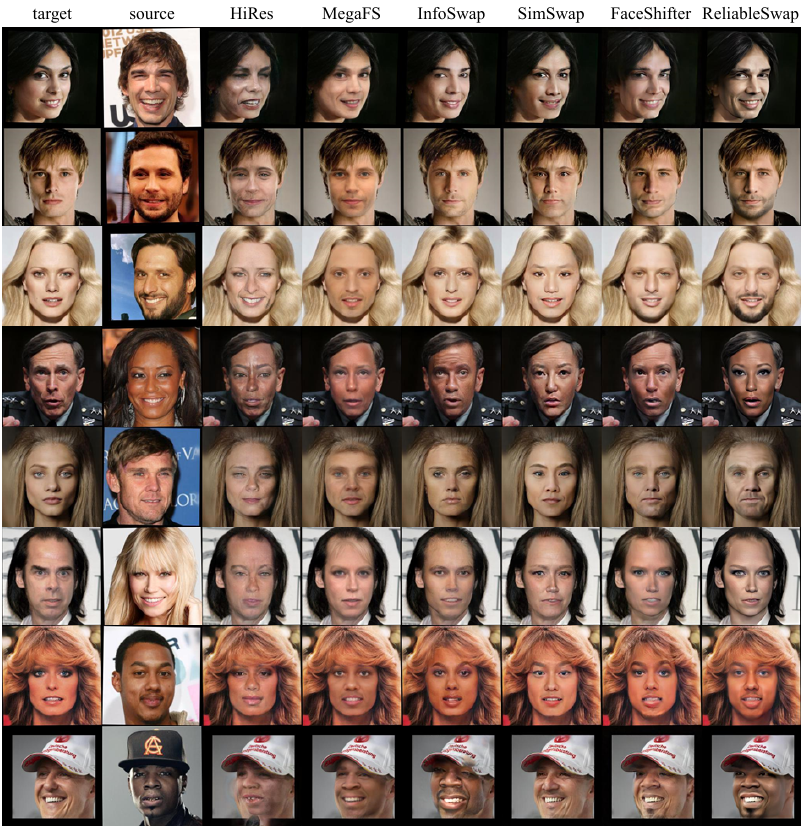}
       \caption{Qualitative comparison on the  CelebA-HQ dataset.}
       \label{fig:supp_celebahq}
    \end{figure*}

    \begin{figure*}[t]
      \centering
       \includegraphics[width=1.0\linewidth]{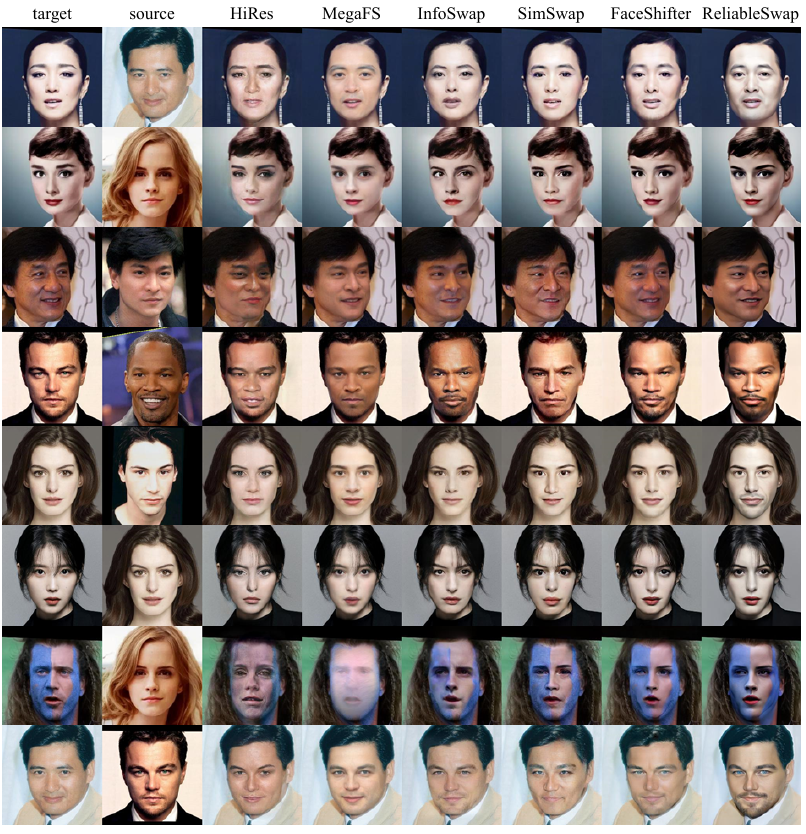}
       \caption{Qualitative comparison on wild faces.}
       \label{fig:supp_web}
    \end{figure*}

We present more qualitative comparison on CelebA-HQ~\cite{karras2017progressive} samples (in Fig.~\ref{fig:supp_celebahq}) and other wild faces (in Fig.~\ref{fig:supp_web}).
The results show that our ReliableSwap (\emph{w/} FaceShifter) achieves better preservation of source identity compared to the other methods.

\section{Additional Experiments on SimSwap}
\label{simswap}

\begin{figure*}[t]
    \centering
    \includegraphics[width=0.95\linewidth]{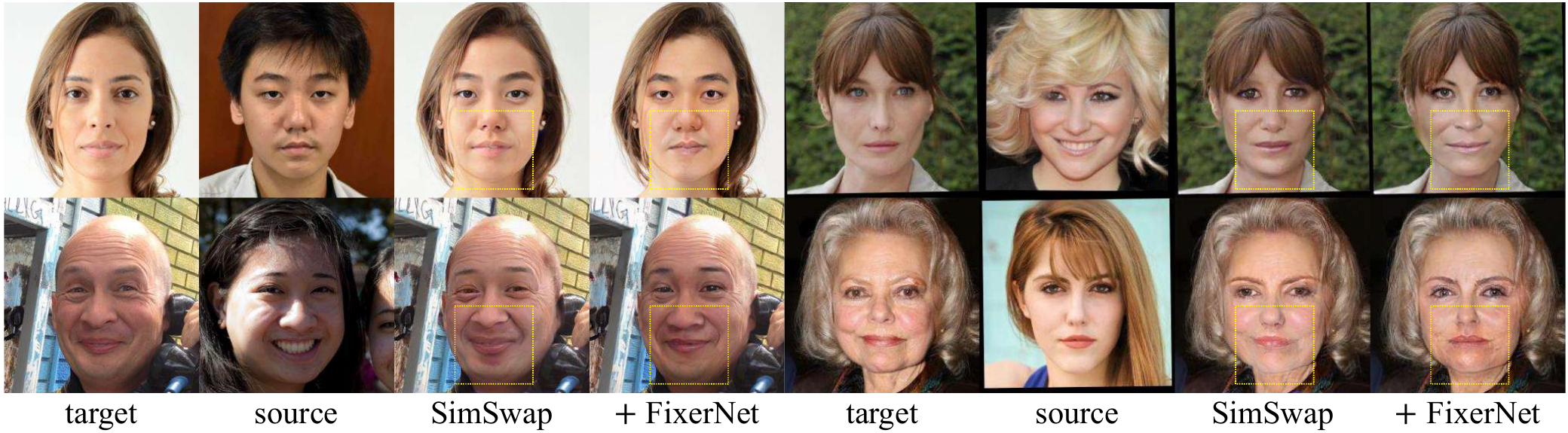}
    \vspace{-0cm}
    \caption{We train FixerNet on the  AsianCeleb dataset and it improves lower face consistency on the SimSwap alone.}
    \vspace{-0cm}
    \label{fig:rebuttal_fixernet}
\end{figure*}

Our FixerNet does not rely on any specific dataset or method.
Any face swapping methods lacking lower face consistency can be enhanced by our FixerNet.
Based on it, our ReliableSwap improves FaceShifter and SimSwap on lower-face consistency (see Fig.~\ref{fig:ffplus}).
Furthermore, we provide Fig~\ref{fig:rebuttal_fixernet} to show the improvement of FixerNet on the SimSwap baseline alone, where FixerNet is trained on AsianCeleb~\cite{asianceleb} dataset.

\begin{figure*}[t]
    \centering
    \includegraphics[width=0.95\linewidth]{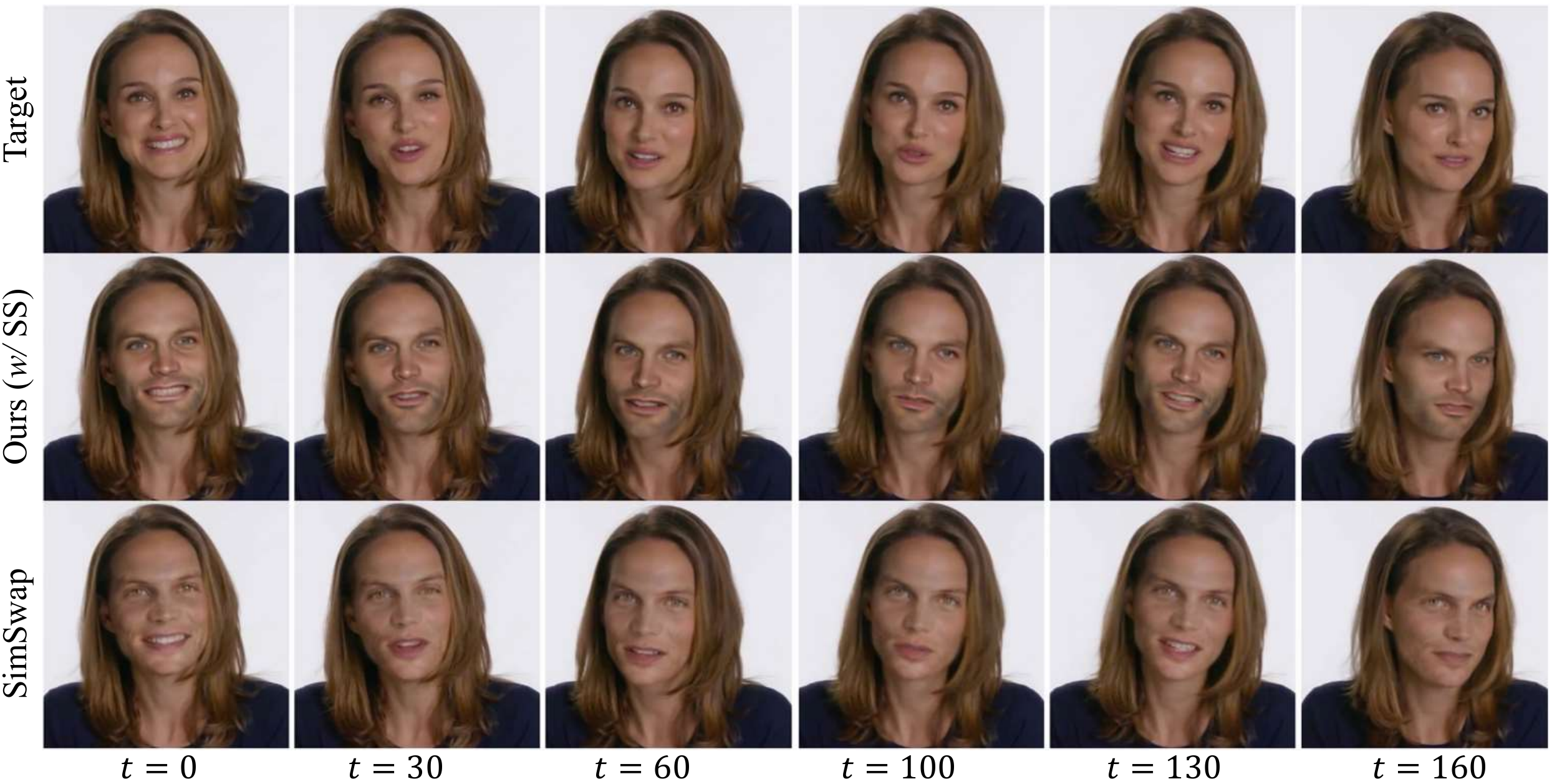}
    \vspace{-0.cm}
    \caption{Video frames of Simswap and ours.}
    \vspace{-0.cm}
    \label{fig:rebuttal_video_comparison}
\end{figure*}

Given the the top-left image in Fig~\ref{fig:rebuttal_styleface} as the source face, Fig~\ref{fig:rebuttal_video_comparison} shows the video frames results of SimSwap and Ours (\textit{w}/ SimSwap), indicating that our method is robust to different camera angles when taking SimSwap as the baseline.
Through using our method, more consistency source identity can be preserved.

\section{Comparisons with StyleFace}
\label{styleface}

\begin{figure*}[t]
    \centering
    \includegraphics[width=0.95\linewidth]{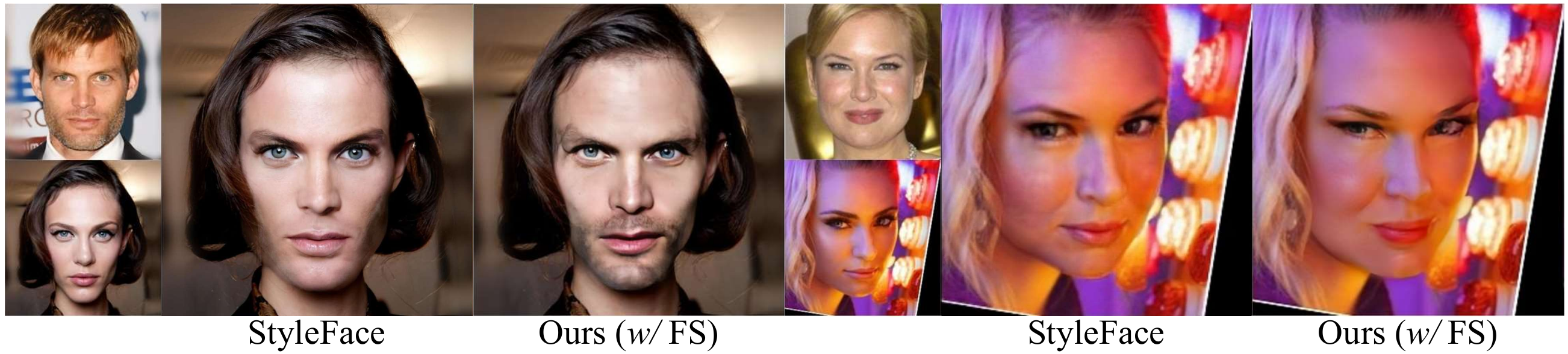}
    \vspace{-0.cm}
    \caption{Comparison between StyleFace and ours.}
    \vspace{-0.cm}
    \label{fig:rebuttal_styleface}
\end{figure*}


Our ReliableSwap provides a novel training scheme for face swapping, which is orthogonal with other methods.
For fairness, we follow the same experimental setting with the baselines. 
In theory, our method can be easily applied to other methods supporting higher resolution.
However, few methods working on higher resolution (512$^2$ or 1024$^2$) provide their training codes.
Nonetheless, we provide the performance of our method (\textit{w}/ FaceShifter) finetuned on 512$^2$ in Fig~\ref{fig:rebuttal_styleface} for reference. Here, to be fair, the StyleFace~\cite{luo2022styleface} results cropped from its paper are resized from 1024 to 512.
Compared with StyleFace, ours preserves better source ID and comparable target attributes.

\section{Face Swapping in Videos}
\label{video}

To evaluate the performance on video face swapping, we randomly choose several video clips from CelebV-HQ~\cite{celebvhq2022eccv} dataset as the target face videos.
For source images, we randomly sample the faces from CelebA-HQ~\cite{karras2017progressive} and the InterNet.
The comparison results are shown in the supplementary file \href{https://www.youtube.com/watch?v=uqe4pD-XpGE}{``demo.mp4''}.

\end{document}